%% file: iccp20_main.tex
\documentclass[10pt,journal,compsoc]{IEEEtran}
\newif\ifpeerreview

\peerreviewfalse

\usepackage{color}
\newcommand{\highlighttext}[1] {{#1}}
\usepackage{multirow}

\usepackage[nocompress]{cite}
\usepackage{url}
\usepackage{amsmath,amssymb,graphicx}

\usepackage{booktabs}

\usepackage[switch]{lineno}

\newcommand{\paperID}{11}

\title{Deep Slow Motion Video Reconstruction with Hybrid Imaging System}

\author{Avinash~Paliwal and~Nima~Khademi~Kalantari
\IEEEcompsocitemizethanks{\IEEEcompsocthanksitem A. Paliwal and N.K. Kalantari are with the Department of Computer Science and Engineering, Texas A\&M University, College Station, TX 77843. E-mail: \{avinashpaliwal, nimak\}@tamu.edu}\protect\\
}

\begin{document}

\IEEEtitleabstractindextext{%
\begin{abstract}
Slow motion videos are becoming increasingly popular, but capturing high-resolution videos at extremely high frame rates requires professional high-speed cameras. To mitigate this problem, current techniques increase the frame rate of standard videos through frame interpolation by assuming linear object motion which is not valid in challenging cases. In this paper, we address this problem using two video streams as input; an auxiliary video with high frame rate and low spatial resolution, providing temporal information, in addition to the standard main video with low frame rate and high spatial resolution. We propose a two-stage deep learning system consisting of alignment and appearance estimation that reconstructs high resolution slow motion video from the hybrid video input. For alignment, we propose to compute flows between the missing frame and two existing frames of the main video by utilizing the content of the auxiliary video frames. For appearance estimation, we propose to combine the warped and auxiliary frames using a context and occlusion aware network. We train our model on synthetically generated hybrid videos and show high-quality results on a variety of test scenes. To demonstrate practicality, we show the performance of our system on two real dual camera setups with small baseline.

\end{abstract}

}

\maketitle
\thispagestyle{empty}

\input{Introduction.tex}
\input{RelatedWork.tex}
\input{Algorithm.tex}
\input{Training.tex}
\input{Results.tex}
\input{Conclusions.tex}

\ifpeerreview \else
\section*{\highlighttext{Acknowledgments}}
\highlighttext{The authors would like to thank the reviewers for their comments and suggestions. We also thank Luke M. Schmidt for providing insight into designing the digital camera rig, and Deepankar Chanda for help with collection of videos. The \textsc{Juggler} and \textsc{Horse} scenes are from YouTube channels {\em Curtis Lahr} and {\em Sony India}, respectively. This work was in part supported by TAMU T3 grant - 246451.}
\fi

\bibliographystyle{IEEEtran}
\bibliography{references}

\vspace{-0.45in}
\begin{IEEEbiography}[{\includegraphics[width=1in,height=1.25in,clip,keepaspectratio]{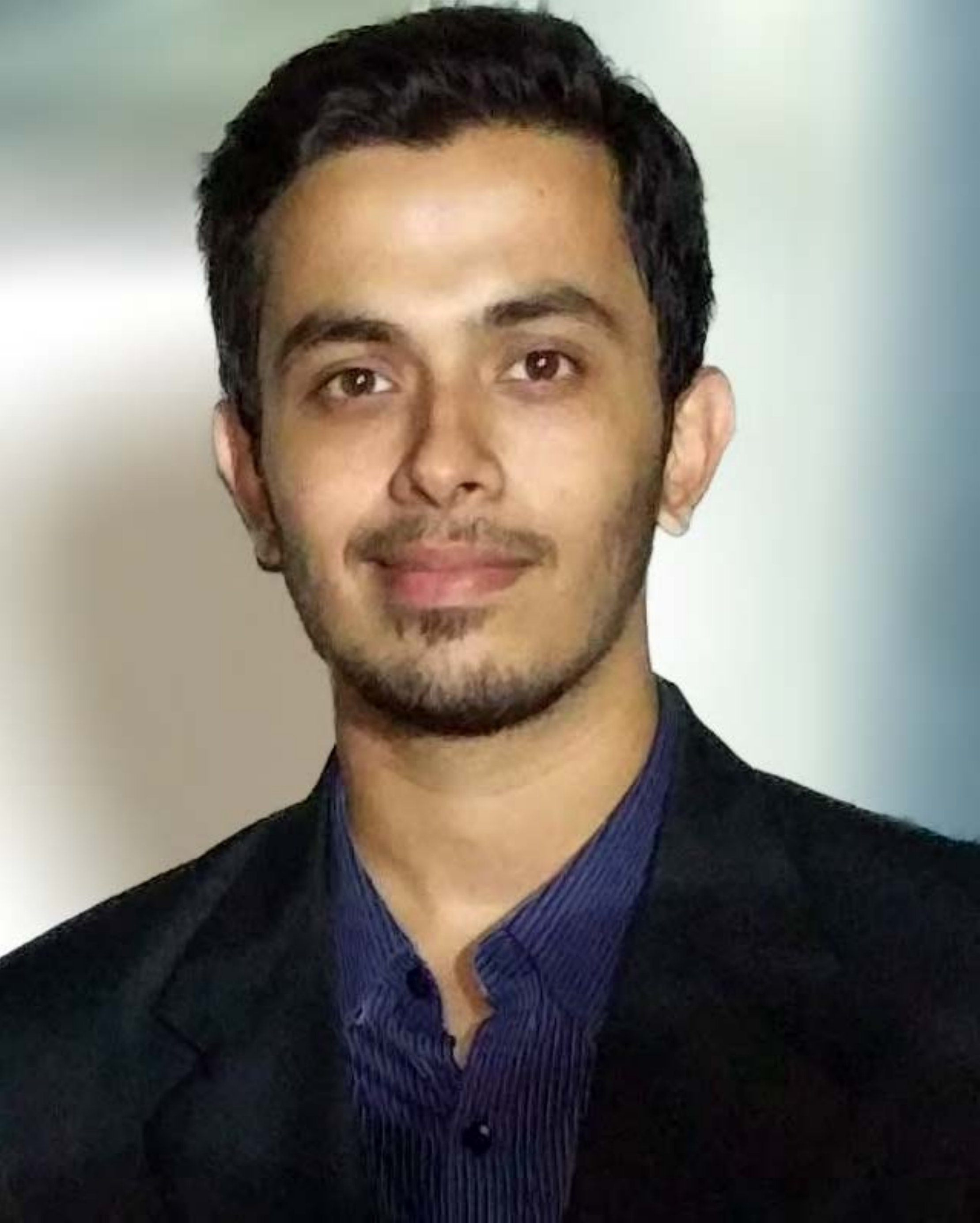}}]{Avinash Paliwal}
is a Ph.D. student at the Computer Science and Engineering Department of Texas A\&M University, College Station, TX, USA. He received his B.Tech. degree in Electronics and Communication Engineering from Visvesvaraya National Institute Of Technology, Nagpur, India. His research interests are in computational photography and computer vision.
\end{IEEEbiography}
\vfill
\vspace{-0.6in}
\begin{IEEEbiography}[{\includegraphics[width=1in,height=1.25in,clip,keepaspectratio]{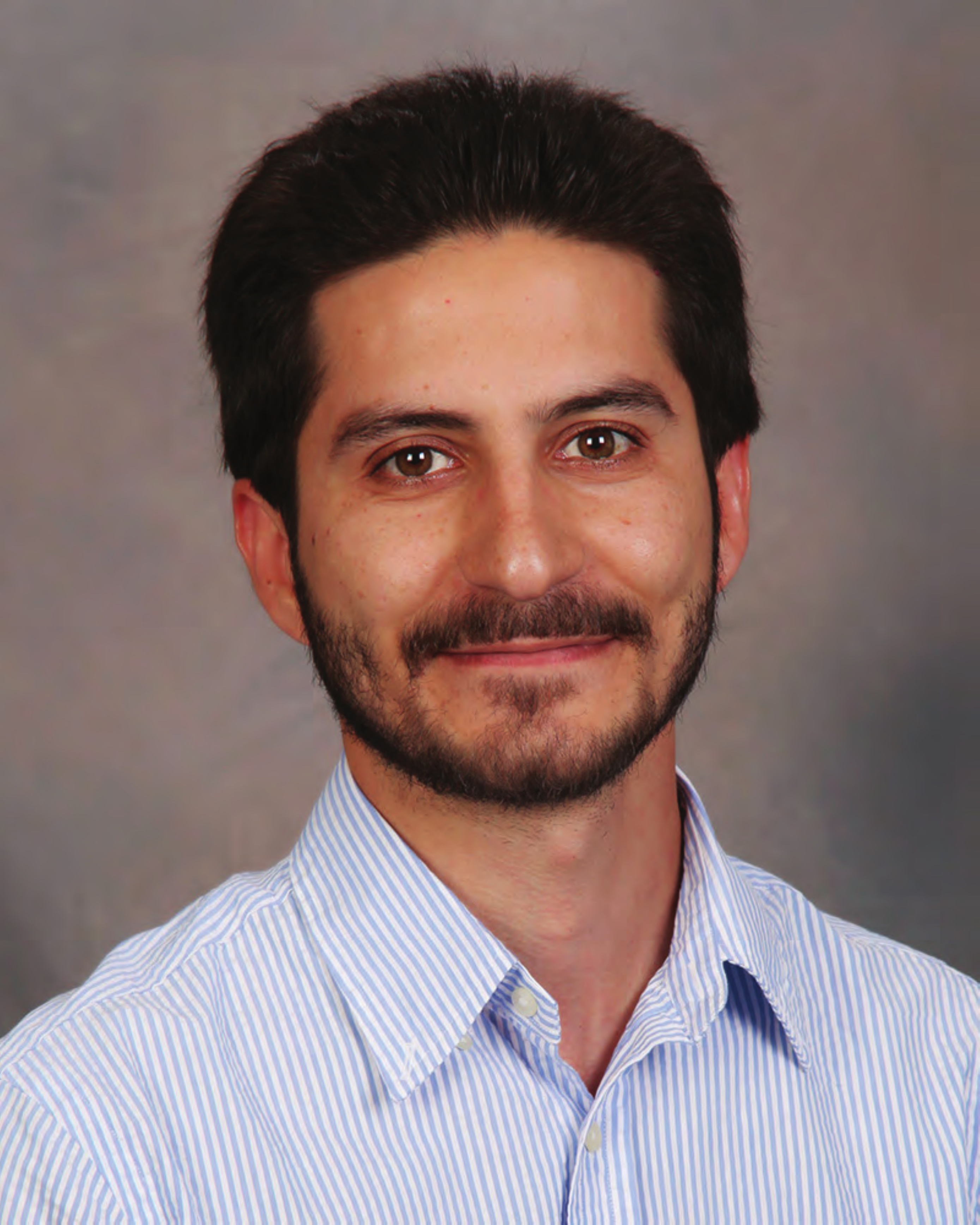}}]{Nima Khademi Kalantari}
is an assistant professor in the Computer Science and Engineering department at Texas A\&M University. He received his doctoral degree in Electrical and Computer Engineering from the University of California, Santa Barbara. Prior to joining Texas A\&M, he was a postdoc in the Computer Science and Engineering department at the University of California, San Diego. His research interests are in computer graphics with an emphasis on computational photography and rendering.
\end{IEEEbiography}




\end{document}

%% file: Introduction.tex
\section{Introduction}
\label{sec:Introduction}

\begin{figure*}
 \includegraphics[width=\linewidth]{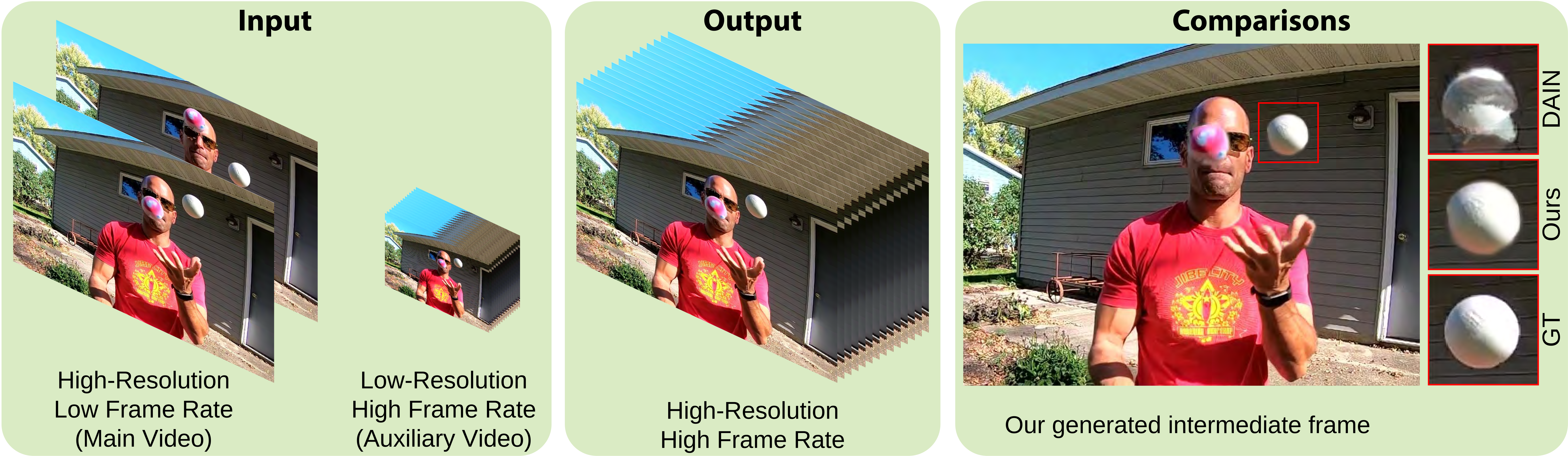}
 \centering
 \vspace{-0.25in}
  \caption{We propose to increase the frame rate of standard videos with low frame rate from hybrid video inputs. In addition to the standard high-resolution video with low frame rate, our system takes a high frame rate video with low resolution as the input (left). We propose a deep learning approach to produce a high-resolution high frame rate video from the two input video streams (shown in the middle). As shown on the right, our approach produces high-quality results that are better than the recent learning-based video interpolation method by Bao et al. (DAIN)~\cite{DAIN}. The complete comparison against other methods is shown in Fig.~\ref{fig:SynResults} and the comparison videos are provided in the supplementary material.}
  \label{fig:Teaser}
  \vspace{-0.18in}
\end{figure*}

Many moments in life, such as popping a bottle of champagne or lightning, happen in a short period of time and are difficult to observe in real time speed. These moments can be properly recorded with a high frame rate camera and the resulting video can be played in slow motion. Nowadays, even smartphones have the ability to capture videos with high frame rates and, thus, slow motion videos are becoming increasingly popular. However, since these cameras have limited bandwidth, they increase the frame rate by sacrificing the spatial resolution. For example, Nokia 6.1 can capture 1080p videos at 30 frames per second (fps), but can only capture 480p videos at 120 fps. Although there are professional high-speed cameras that are able to capture high-resolution videos at extremely high frame rates, they are expensive and, thus, not available to the general public.

\begin{figure}
  \includegraphics[width=\linewidth]{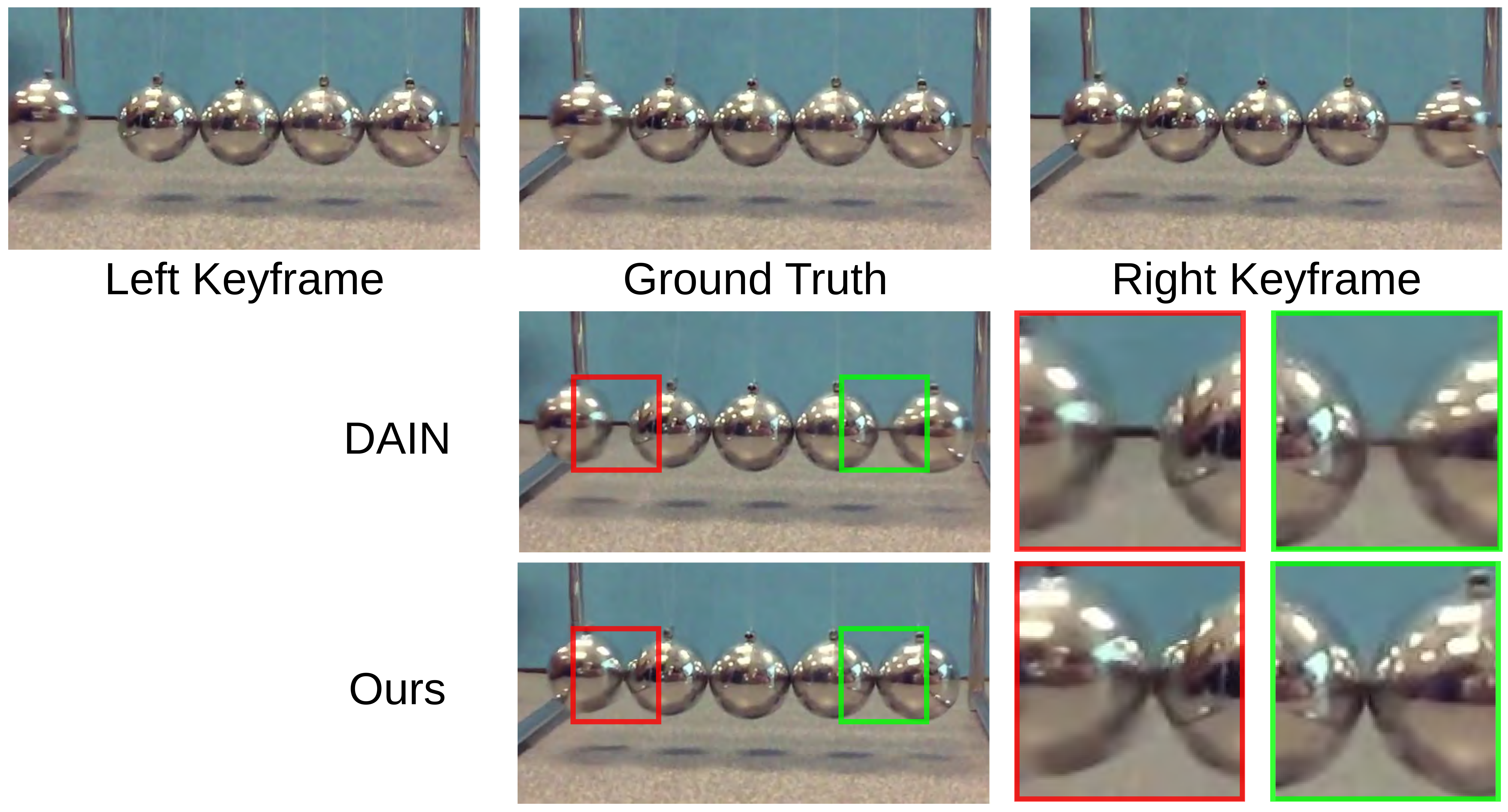}
  \vspace{-0.25in}
  \caption{The top row shows three frames of a video demonstrating Newton's cradle in motion. The state-of-the-art video interpolation techniques, such as the method by Bao et al. (DAIN)~\cite{DAIN}, attempt to interpolate the intermediate frame using the neighboring keyframes by assuming the motion between them to be linear. However, this is not a valid assumption in this case. For example, the rightmost ball stays still in the left and middle frames of the video and then moves to the right and, thus, has a non-linear motion. Since Bao et al.'s approach linearly interpolates the motion of the left and right balls, it produces incorrect results, as shown in the red and green insets. Our hybrid system utilizes the temporal information of the additional high frame rate video with low spatial resolution and is able to properly handle this challenging case.}
  \label{fig:TemporalAliasing}
  \vspace{-0.20in}
\end{figure}

To address this problem, existing techniques use a standard high-resolution video at low frame rate as input and increase the frame rate by interpolating the frames in between~\cite{Liu17,Niklaus17_ICCV,Jiang18,Niklaus18,liu2019cyclicgen,DAIN}. Although these methods produce high-quality results for simple cases, their ability to handle challenging cases with complex motion is limited. This is mainly because all of them have a limiting assumption that the motion between the two existing frames is linear. However, this is not the case in challenging scenarios, as shown in Fig.~\ref{fig:TemporalAliasing}. 

In this paper, we address this problem by using a set of hybrid videos consisting of a {\em main} and an {\em auxiliary} video captured with a small baseline. As shown in Fig. 1, the main video has high resolution and low frame rate, while the auxiliary video is of low resolution and high frame rate. Therefore, the main and auxiliary frames provide the spatial and temporal information, respectively. The goal of our approach is to use the main and auxiliary videos as input and produce a video with high spatial resolution (same as the main video) and high frame rate (equal to the auxiliary video). This setup takes advantage of the current trend of dual cameras in the smartphones.

We propose to reconstruct the missing target frames in the main video through a learning-based system consisting of two main stages of alignment and appearance estimation. During alignment, we use a pre-trained flow estimation CNN to compute initial flows between the target and neighboring frames using only the auxiliary frames. We then use a CNN to enhance these estimates by incorporating the information in the high resolution main frames and utilize the enhanced flows to warp the main frames. In the next stage of appearance estimation, we propose a context and occlusion aware CNN to combine the warped and target auxiliary frames and produce the final image.

We train our system in an end-to-end fashion by minimizing the loss between the synthesized and ground truth frames on a set of training videos. For training data, we use a set of high frame rate videos with high resolution as the ground truth and propose an approach to synthetically generate the main and auxiliary input videos, emulating the inputs from two real cameras with a small baseline. To demonstrate our results on real scenarios, we build two setups; a camera rig consisting of two smartphones and another one with two digital cameras. Experimental results shows that our approach is able to produce high-quality results on both synthetic and real datasets.

In summary, the contributions of this work are:
\vspace{-0.02in}
\begin{itemize}
    \item We propose a two-stage deep learning model that combines two input video streams from a hybrid imaging system to reconstruct a high quality slow motion video.
    \item We present a flow estimation system that utilizes two videos to generate high resolution flows even in challenging cases with large motion.
    \item We introduce a context and occlusion aware appearance estimation network which blends the two warped key frames and minimizes warping artifacts.
    \item We construct two real dual camera rigs with small baseline to demonstrate the results of our model on real world data.
\end{itemize}

%% file: RelatedWork.tex
\vspace{-0.2in}
\section{Related Work}
\label{sec:RelatedWork}

Frame interpolation is a classical problem in computer graphics with applications like increasing the frame rate of videos and novel view synthesis. This problem has been extensively studied in the past and many powerful methods have been developed. Here, we focus on the approaches performing video frame interpolation \highlighttext{and optical flow estimation, a common ingredient in existing video frame interpolation systems}. We also briefly discuss the approaches using hybrid \highlighttext{imaging} systems.

\vspace{-0.05in}
\subsection{\highlighttext{Optical Flow}} 
\highlighttext{
Optical flow estimation has been a long standing problem in the field of computer vision with a plethora of papers trying to address the challenges like large motion and occlusion. Many classical approaches~\cite{memin1998dense, brox2004high, wedel2009structure} are based on Horn and Schunck's~\cite{horn1981determining} seminal approach which minimized a custom energy function based on brightness constancy and spatial smoothness assumptions. Sun et al.~\cite{sun2014quantitative} provides a comprehensive comparison between these approaches. These methods however require solving complex optimization problems and in most cases are computationally expensive.}

\highlighttext{
Most of the recent state-of-the-art approaches have utilized deep learning for the flow estimation problem. The FlowNet models by Dosovitskiy et al.~\cite{long2015fully} are the first end-to-end trained deep CNNs for estimating optical flow. Ilg et al.~\cite{ilg2017flownet} further improve this approach by estimating large and small displacements through a series of FlowNet models and then fusing the results together. Inspired by the classical coarse-to-fine pyramid approach for image registration, Ranjan
and Black~\cite{ranjan2017optical} introduce a spatial pyramid network to estimate flow at a coarse scale and then refine this flow by computing residuals at higher resolutions. Similarly, Sun et al.~\cite{Sun18} propose a pyramid network which utilizes features instead of images. This architecture has the advantage of being fast and compact while also giving a better performance. We utilize this model for estimating an initial flow between auxiliary frames since it works well with high resolution videos containing large motion.
}

\highlighttext{Another popular approach is to train a deep CNN using unsupervised learning~\cite{Yin2018GeoNet, yu2016back, ren2017unsupervised, wang2018occlusion}. Instead of using ground truth flows, the generated flows are used as input for a secondary task and the output quality is used as a supervisory signal to train the flow network. For example, Long et al.~\cite{long2016learning} supervise their optical flow CNNs by interpolating frames. We utilize a similar strategy to train our high resolution flow estimation network.}

\begin{figure}
  \includegraphics[width=\linewidth]{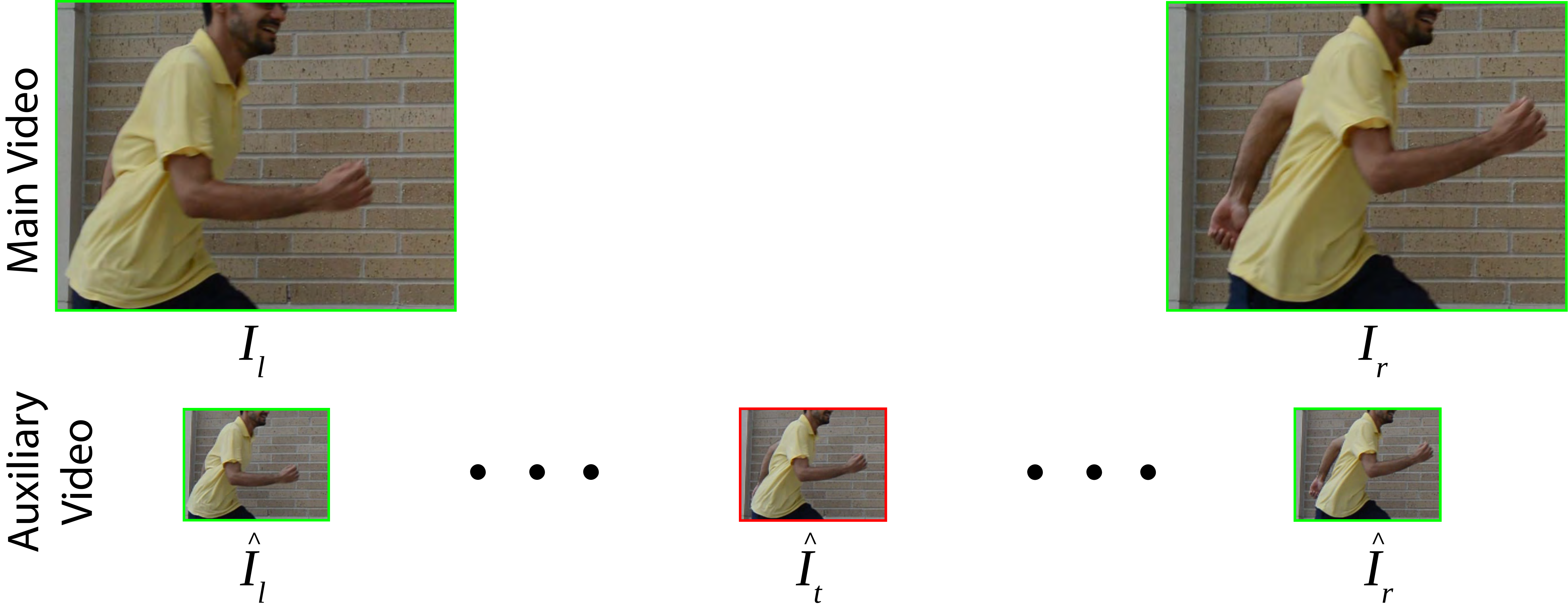}
  \vspace{-0.32in}
  \caption{Our system takes two video streams with different spatial resolutions and frame rates as the input. On the top, we show the frames from our main video with high-resolution and low frame rate. The auxiliary frames with low resolution and high frame rate are shown at the bottom.}
  \label{fig:Inputs}
  \vspace{-0.1in}
\end{figure}

\begin{table}[!t]
\renewcommand{\arraystretch}{1.3}
\caption{Notations used in the paper}
\vspace{-0.15in}
\centering
\begin{tabular}{cl}
  \hline\hline
  $t$     & index of the target frame\\
  $l$     & index of the keyframe {\em before} the target frame\\
  $r$     & index of the keyframe {\em after} the target frame\\
  $I_n$   & frame $n$ of the {\em main} video\\
  $\hat{I}_n$   & frame $n$ of the {\em auxiliary} video\\
  $F_{n}$ & flow from the target frame $I_t$ to $I_n$ in the {\em main} video\\
  $\hat{F}_{n}$ & flow from the target frame $\hat{I}_t$ to $\hat{I}_n$ in the {\em auxiliary} video\\
  $V_{l}$ & visibility map of the warped left keyframe. $V_r$ is\\ & computed as: $V_r = 1-V_l$.\\
  \hline\hline
\end{tabular}
\vspace{-0.15in}
\label{tab:Notations}
\end{table}

\vspace{-0.05in}
\subsection{Video Frame Interpolation}

A standard solution to this problem is to compute optical flows~\cite{Baker11} between the neighboring frames and use the estimated flows to warp the existing frames and reconstruct the missing frame. However, optical flow is brittle in presence of large motion and also has difficulty handling the occluded regions. Mahajan et al.~\cite{Mahajan09} propose to move and copy the gradients across each pixels path. They then perform Poisson reconstruction on the interpolated gradients to obtain the final image. \highlighttext{Shechtman et al.~\cite{shechtman2010regenerative} use patch-based synthesis for reconstructing the intermediate frames.} Meyer et al.~\cite{Meyer15} propose to interpolate the phase across different levels of a multi-scale pyramid.

In recent years, several approaches have utilized deep learning for this application. Niklaus et al.~\cite{Niklaus17_CVPR} use a deep CNN to estimate a spatially adaptive convolutional kernel for each pixel. This kernel is then used to reconstruct the final image from the two neighboring frames. Later, they improved the speed and quality of the results by formulating the problem as local separable convolution and using a perceptual loss~\cite{Niklaus17_ICCV}. Liu et al.~\cite{Liu17} use a CNN to estimate the flows and use them to reconstruct the missing frame by trilinear interpolation.

Niklaus and Liu~\cite{Niklaus18} propose to blend the pre-warped neighboring frames using a CNN. They also utilize contextual information in the blending stage to improve the quality of the results. Meyer et al.~\cite{Meyer18} improves the phase-based approach~\cite{Meyer15} by using a CNN to estimate the phase of the intermediate frames across different levels of a pyramid. Jiang et al.~\cite{Jiang18} propose to interpolate arbitrary number of in between frames by estimating flows to the intermediate frame. The flows are then used to warp the neighboring frames which are in turn combined to generate the final frame using a weighted average. Liu et al.~\cite{liu2019cyclicgen} propose that the synthesized interpolated frames are more reliable if they are cyclically used to reconstruct the input frames. Wenbo et al.~\cite{DAIN} introduce a depth-aware interpolation approach to better handle occlusions in the video frames.

All of these approaches work with the main assumption that the motion between the existing frames is linear, but this is not a valid assumption in challenging cases. Therefore, they are not able to correctly synthesize the intermediate frames in cases with complex motion, as shown in Fig.~\ref{fig:TemporalAliasing}. In our system, we use an additional auxiliary video to provide the needed temporal information and, thus, are able to produce high-quality results without aliasing.

\begin{figure}
 \includegraphics[width=\linewidth]{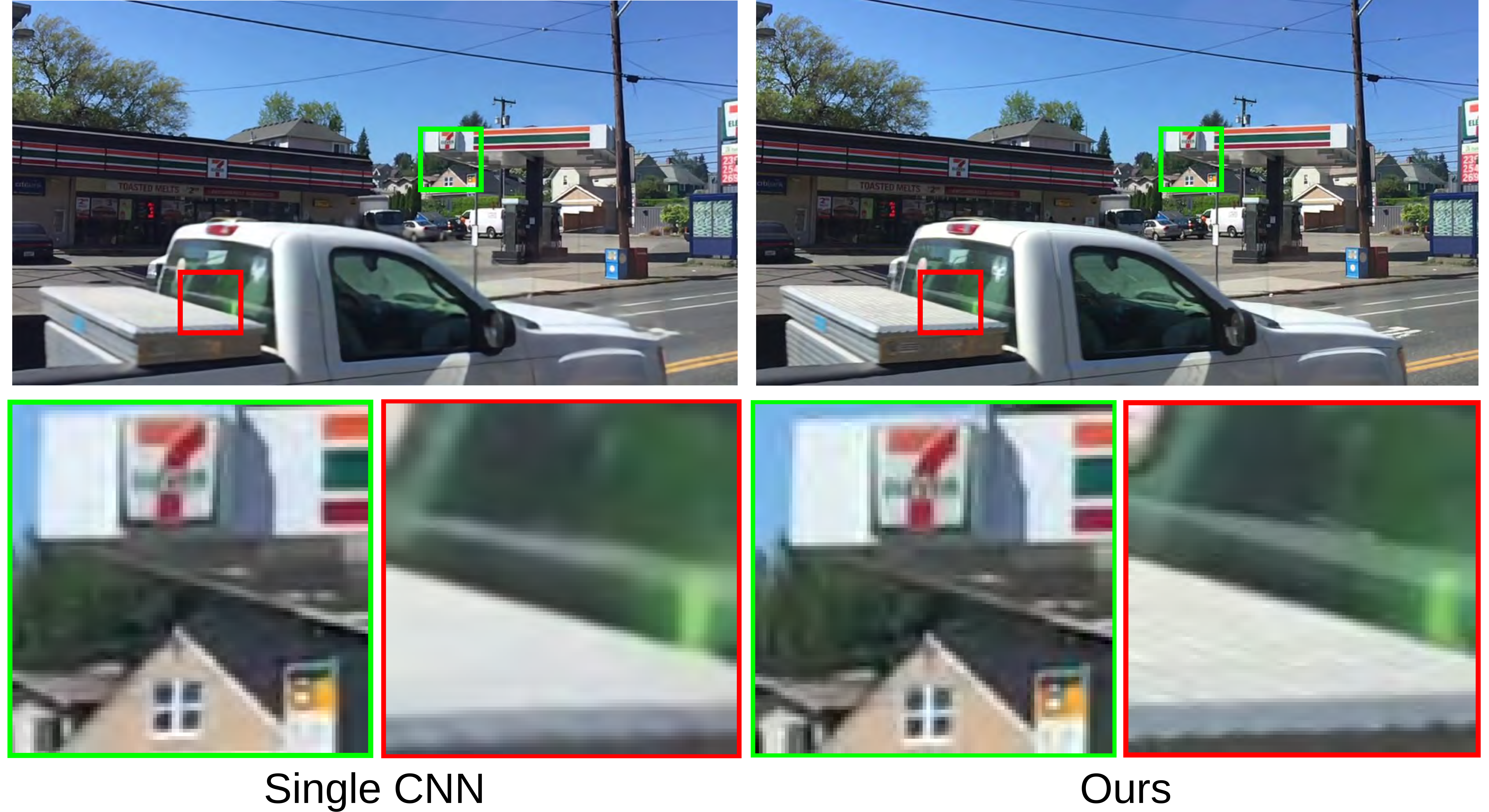}
 \vspace{-0.30in}
 \caption{We compare the result of our two-stage approach against the simpler single stage method where a single network directly reconstructs the high-resolution target frame from the two keyframes and the corresponding auxiliary target frame. The single CNN quickly learns to combine the keyframes with the auxiliary frame without properly aligning them. Therefore, it can produce reasonable results in the static regions, as shown in the green inset. However, in the regions with motion, like the one shown in the red inset, it produces result with excessive blurriness. Our method, on the other hand, properly aligns the keyframes and is able to produce high-quality results in both regions.}
 \label{fig:Straightforward}
  \vspace{-0.15in}
\end{figure}

\vspace{-0.1in}
\subsection{Hybrid \highlighttext{Imaging} System} Several approaches have used multiple cameras to increase the spatial resolution of images~\cite{Sawhney01,Bhat07,Wang16_CVPR} and both spacetime resolution of videos~\cite{gupta:2009, Watanabe}. Boominathan et al.~\cite{Boominathan14} and Wang et al.~\cite{Wang17} combine a DSLR camera with a light field camera to increase the resolution of light fields spatially and temporally, respectively. \highlighttext{Ben-Ezra and Nayar~\cite{ben-ezra2004motion} construct a hybrid imaging system to estimate camera motion and use it to deblur the captured image.}

A few non-learning methods~\cite{Watanabe, gupta:2009} have proposed to synthesize high resolution high frame rate videos using a dual camera setup. Watanabe et al.~\cite{Watanabe} first estimate motion in the video and then perform fusion in the wavelet domain to synthesize the final frame. Gupta et al.~\cite{gupta:2009} compute flows using the high resolution stills and low resolution video, and then perform graph-cut based fusion to generate the output. Their fusion algorithm tries to simultaneously preserve the spatial and temporal gradients, essentially creating a trade-off between spatial resolution and temporal consistency. They give more weight to temporal consistency which results in blurry patches in the output video for challenging scenes. In contrast, our learning based technique is able to effectively combine the information of the two input videos and produce high-quality videos, as shown in Fig.~\ref{fig:GuptaComp}.

%% file: Algorithm.tex
\begin{figure*}
  \includegraphics[width=\linewidth]{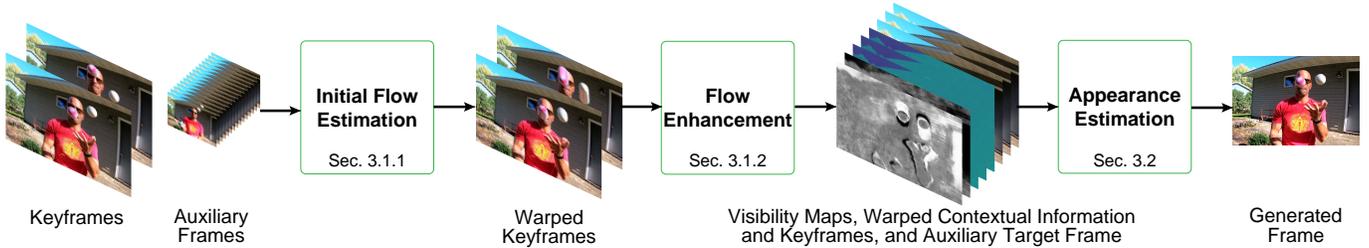}
  \vspace{-0.3in}
  \caption{\textbf{System overview}: We first estimate low resolution flows from the target frame using auxiliary frames. These flows are then enhanced by using the high resolution keyframes and warped keyframes in the flow enhancement network. The final warped frames generated with the enhanced flows are masked with the visibility maps and then combined by the context-aware appearance estimation network to output the final frame.}
  \label{fig:ArchOvw}
  \vspace{-0.15in}
\end{figure*}

\section{Algorithm}
\label{sec:Algorithm}

The inputs to our system are two video streams with different spatial resolutions and frame rates. In our system, the {\em main} low frame rate video captures high spatial resolution details, as shown in Fig.~\ref{fig:Inputs}. The {\em auxiliary} video, on the other hand, is a high frame rate video with low spatial resolution. For example, our digital camera system (shown on right in Fig.~\ref{fig:Rig}) captures the main video with resolution of $1920\times 720$ at 30 fps, while the auxiliary camera records a video with a resolution $640\times 240$ at 400 fps. Our goal is to use these two video streams as input and reconstruct a video with the spatial resolution and frame rate of the main and auxiliary videos, respectively. To reconstruct each high-resolution target frame, $I_{t}$, we need to utilize the information in the two neighboring frames in the main video (called {\em keyframes} hereafter), $I_{l}$ and $I_{r}$, and the target low resolution frame from the auxiliary video, $\hat{I}_{t}$ (see Fig.~\ref{fig:Inputs}). Table~\ref{tab:Notations} summarizes the notations used in the paper.

A straightforward approach for addressing this problem is to pass the two main keyframes and the bilinearly upsampled auxiliary target frame, $I_{l}$, $I_{r}$, and $\hat{I}_{t}$, as inputs to a CNN and estimate the high-resolution target frame, $I_{t}$. Ideally, the network increases the spatial resolution of $\hat{I}_{t}$ by utilizing the corresponding content from the two keyframes. However, as shown in Fig.~\ref{fig:Straightforward}, this approach produces results with blurriness, specifically in the regions with large motion. This demonstrates that while the network quickly learns to combine the input images to improve the regions with small or no motion, it is unable to align the keyframes in regions with large motion.

Therefore, we propose an end-to-end trained two-stage learning system, consisting of {\em alignment} and {\em appearance estimation}, to force the system to properly learn to align the keyframes and combine them. During alignment, we first use homography to globally align the main and auxiliary frames since they are captured from different views. We then register the high-resolution keyframes to the target frame by estimating a set of flows between them. Next, we combine the aligned keyframes and the auxiliary target frame to produce the final frame in the appearance estimation stage. An overview of the process is shown in Fig.~\ref{fig:ArchOvw}. We discuss these two stages in the following sections.

\subsection{Alignment}
\label{ssec:Alignment}

In order to align the two keyframes to the target frame, we first need to compute a set of flows from the left and right keyframes to the target frame, denoted by $F_l$ and $F_r$, respectively. Once these flows are obtained, we can simply use them to backward warp the two keyframes and produce a set of aligned keyframes, $g(I_l,F_l)$ and $g(I_r,F_r)$, where $g$ is the function that performs backward warping using bilinear interpolation.

Direct estimation of the high-resolution flows, $F_l$ and $F_r$, is difficult as the high-resolution target frame, $I_t$, is not available. Previous methods compute bidirectional flows between neighboring frames and then apply a linear equation to estimate the flow to an intermediate frame assuming linear motion \cite{Jiang18, Niklaus18}. Instead, we propose a novel approach to compute these flows by utilizing the information in both the main and auxiliary frames in two stages. In the first stage, we estimate low resolution flows using the auxiliary frames. We then enhance these flows in the second stage using the high-resolution keyframes.

\begin{figure}
\includegraphics[width=\linewidth]{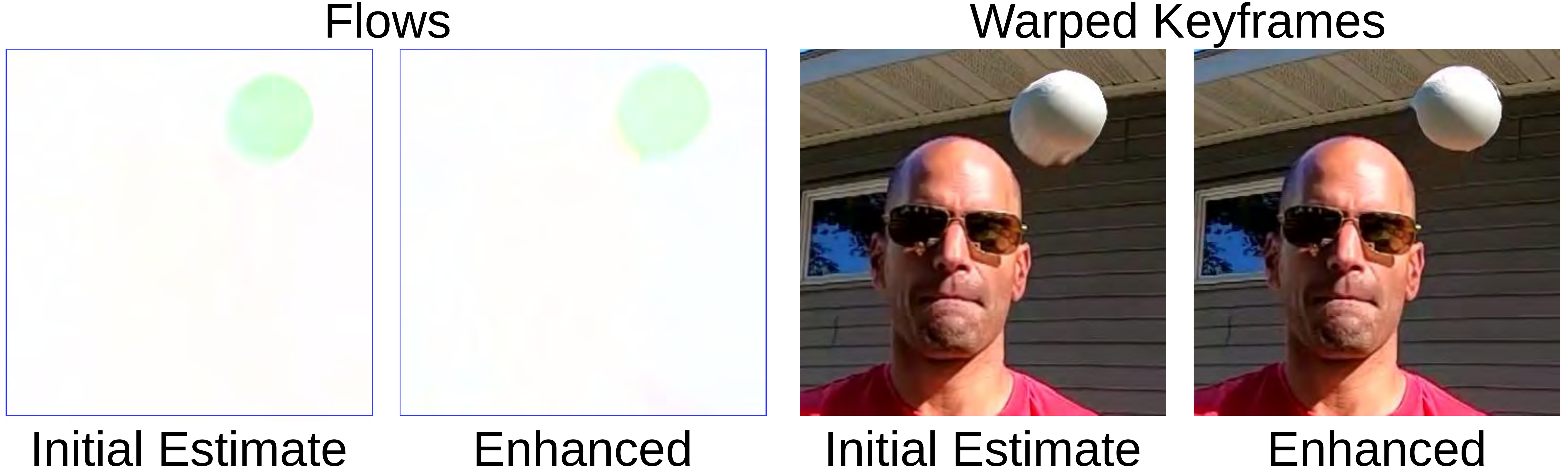}
 \vspace{-0.3in}
\caption{We show the initial and enhanced flows as well as the warped keyframe using these two flows. Since the initial flow is computed using low resolution auxiliary frames, it does not accurately capture the motion boundaries and produces artifacts around the moving ball. The enhanced flow is obtained by utilizing the content of the main frames and, thus, produces warped frame without noticeable artifacts.}
 \vspace{-0.15in}
\label{fig:Flows}
\end{figure}

\subsubsection{Initial Flow Estimation} The goal of this stage is to estimate a set of flows from the two keyframes to the target using the low-resolution auxiliary frames. This can be done by simply computing the flows from $\hat{I}_l$ and $\hat{I}_r$ to $\hat{I}_t$ to obtain $\hat{F}_l$ and $\hat{F}_r$, respectively. However, we observed that in some cases there are large motions between the neighboring and the target frames and, thus, this approach has difficulty estimating the correct flows. Therefore, since we have access to all the intermediate low resolution frames, we propose to compute the flows by concatenating the flows computed between the consecutive auxiliary frames.
\label{sec:InitFlow}
Here, we explain the process for computing $\hat{F}_l$, but $\hat{F}_r$ is computed in a similar manner. To do so, we use PWC-Net by Sun et al.~\cite{Sun18} to compute a set of flows between the consecutive auxiliary frames, i.e., $\hat{I}_{t}$ to $\hat{I}_{t-1}$, $\hat{I}_{t-1}$ to $\hat{I}_{t-2}$, ..., $\hat{I}_{l+1}$ to $\hat{I}_l$. These flows are then \highlighttext{chained} to produce the flow between the left keyframe and the target frame, $\hat{F}_l$. Note that, the resolution of the estimated flows should be equal to the resolution of the main frames, since they are used to warp the main keyframes. We can do this either by computing the low resolution flows and then upsampling them, or first upsampling the low resolution frames and then computing the flows between them. We choose the latter as it generally produces results with higher quality.

While these flows can be used to produce a set of warped keyframes, $g(I_l,\hat{F}_l)$ and $g(I_r,\hat{F}_r)$, they have two main problems. First, these flows are computed on low resolution frames and, thus, are missing high frequency details. Second, while we reduce the disparity between the auxiliary and main frames through homography, there is still minor misalignment between the two videos. Therefore, the flows are often inaccurate around the edges and occlusion boundaries, as shown in Fig.~\ref{fig:Flows}. We improve these flows in the next stage by utilizing the content of the high-resolution main frames.

\begin{figure}
\includegraphics[width=\linewidth]{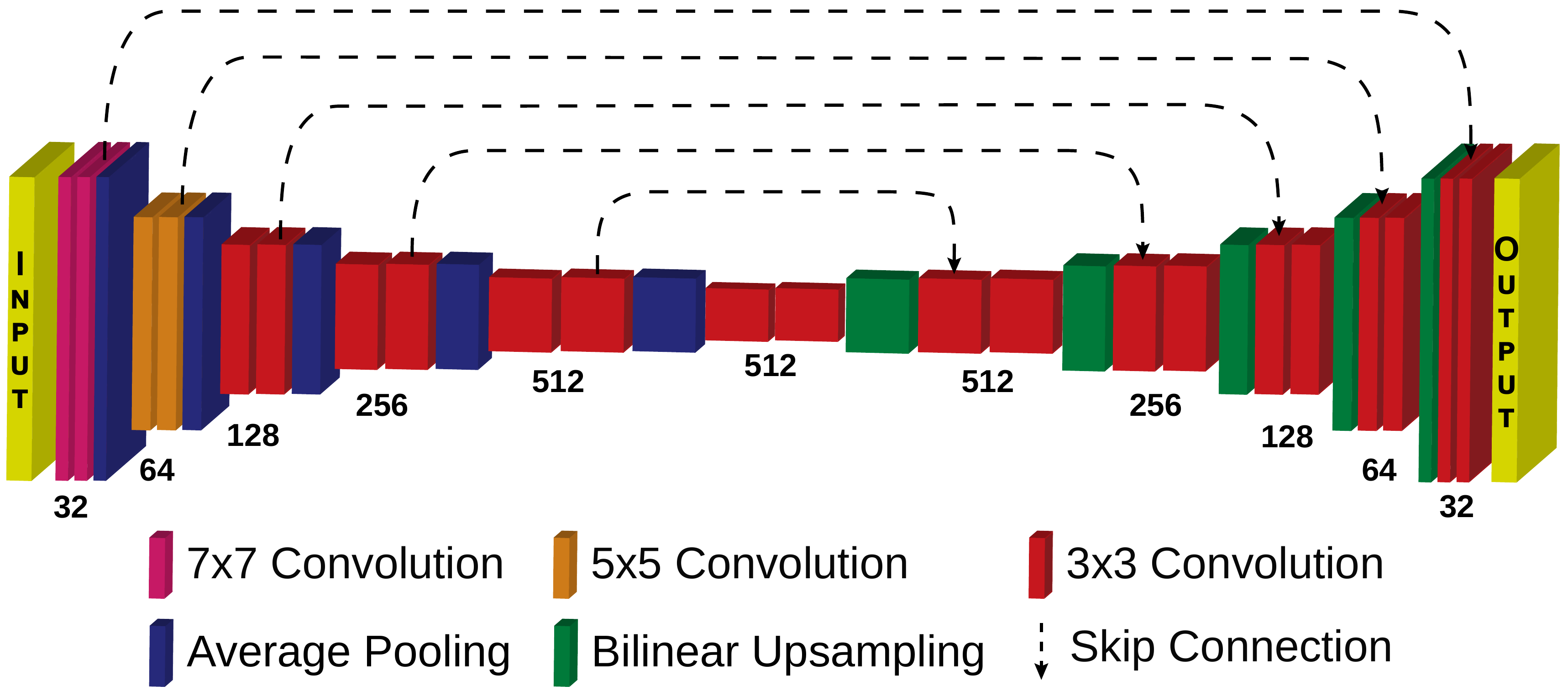}
\vspace{-0.3in}
\caption{Architecture of Flow Enhancement and Appearance Estimation networks. The only difference is inputs and outputs to both networks.}
\label{fig:Arch}
\vspace{-0.2in}
\end{figure}

\subsubsection{Flow Enhancement} Here, the goal is to estimate the residual flows, $\Delta {F}_l$ and $\Delta {F}_r$, that provide high frequency details and alignment for the estimated low resolution flows from the previous stage. The final high resolution flows can then be obtained as:

\vspace{-0.1in}
\begin{equation}
\label{eq:FlowEnhancement}
{F}_l = \hat{F}_l + \Delta {F}_l, \quad {F}_r = \hat{F}_r + \Delta {F}_r
\end{equation}

We propose to estimate the two residual flows using a CNN. The network takes a 19 channel input consisting of the initial flows (2 channels each), $\hat{F}_l$ and $\hat{F}_r$, original and warped keyframes (3 channels each), $I_l$, $I_r$, $g(I_l, \hat{F}_l)$, and $g(I_r, \hat{F}_r)$, as well as the bilinearly upsampled low-resolution target frame (3 channels), $\hat{I}_t$. The estimated residual flows, $\Delta {F}_l$ and $\Delta {F}_r$, can then be used to reconstruct the final flows, using Eq.~\ref{eq:FlowEnhancement}, and they in turn can be used to produce the warped keyframes, $g(I_l, {F}_l)$ and $g(I_r, F_r)$.

Note that, each pixel of the target frame is not always visible in both keyframes due of occlusion. To take this into account, in addition to the residual flows, our flow enhancement network estimates a visibility map for one of the keyframes. In our implementation, we estimate the map for the left keyframe, $V_l$, and obtain the map for the right keyframe as $V_r = 1 - V_l$. These visibility maps are single channel maps and basically measure the quality of the warped image at each pixel. Intuitively, if a region is occluded in a keyframe, the visibility map corresponding to that region will have small values. These visibility maps are extremely helpful in reconstructing the motion boundaries in the appearance estimation stage (see Fig.~\ref{fig:VisibilityEffect}).

For our model, we use an encoder-decoder U-Net architecture \cite{ronneberger2015u, Jiang18} with skip connections, as shown in Fig.~\ref{fig:Arch}. The network is fully convolutional with leaky ReLu (alpha=0.1) activation function. In addition to this, to constrain the range of values in the visibility masks between [0, 1], we use a sigmoid activation function on the output channels corresponding to the visibility map.

\subsection{Appearance Estimation}
\label{ssec:Appearance}

\begin{figure}
 \includegraphics[width=\linewidth]{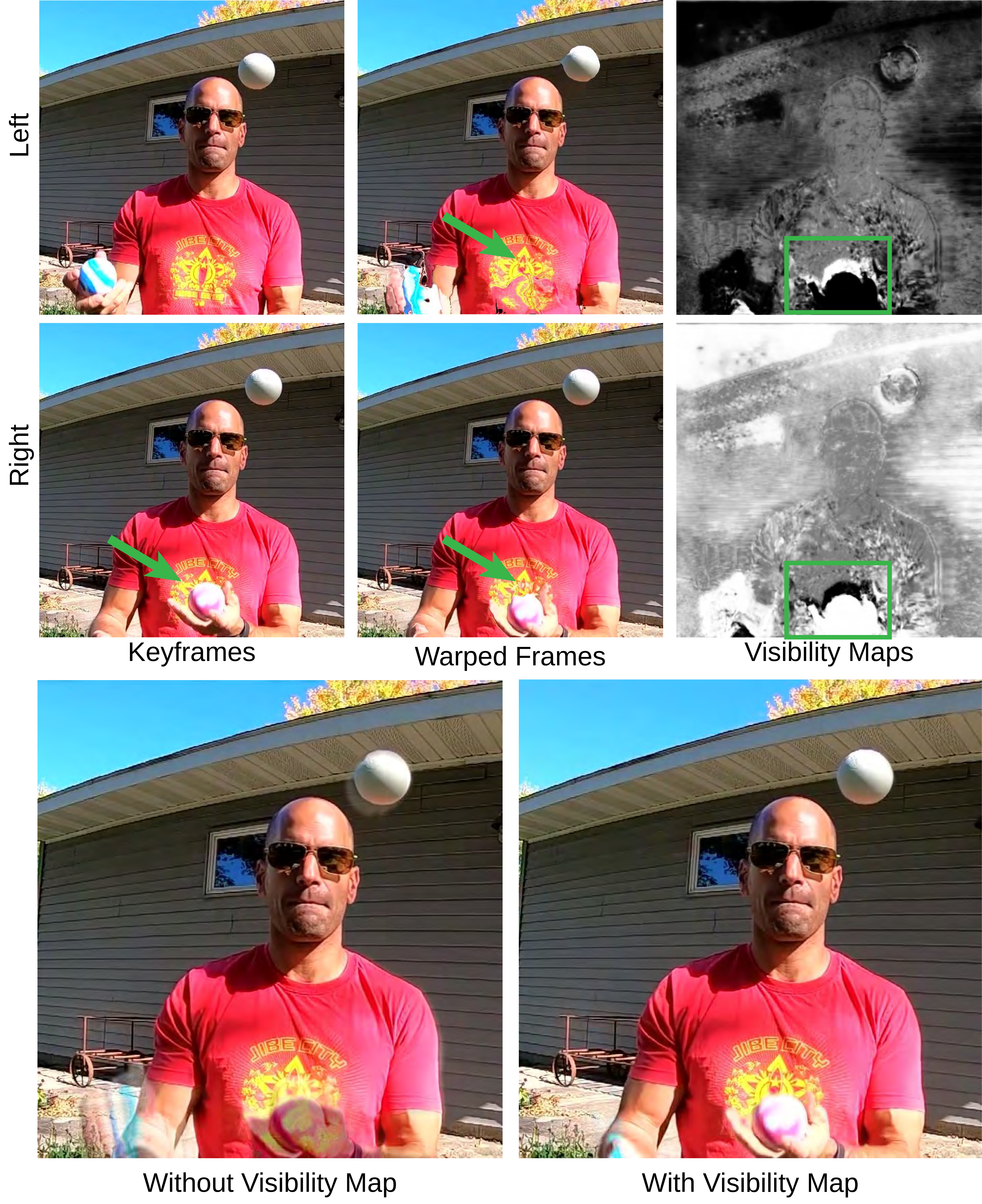}
 \vspace{-0.3in}
 \caption{On the top, we show two keyframes along with their warped version and visibility maps. The juggler's arm is moving up in the video and it only appears in the right keyframe (see the green arrow). Therefore, the warped left keyframe lacks details on the hand since it does not appear in the corresponding keyframe. On the other hand, the warped right keyframe does not have the details on top of the ball, since it is occluded in the right keyframe (note the star on the shirt as indicated by the green arrows). The visibility maps correctly identify the occluded regions and assign lower values to them (see the green boxes). As shown on the bottom, providing the visibility maps to the appearance estimation network, helps utilizing the valid content of the keyframes and producing high-quality results.}
 \label{fig:VisibilityEffect}
 \vspace{-0.12in}
\end{figure}

\begin{figure}
 \includegraphics[width=\linewidth]{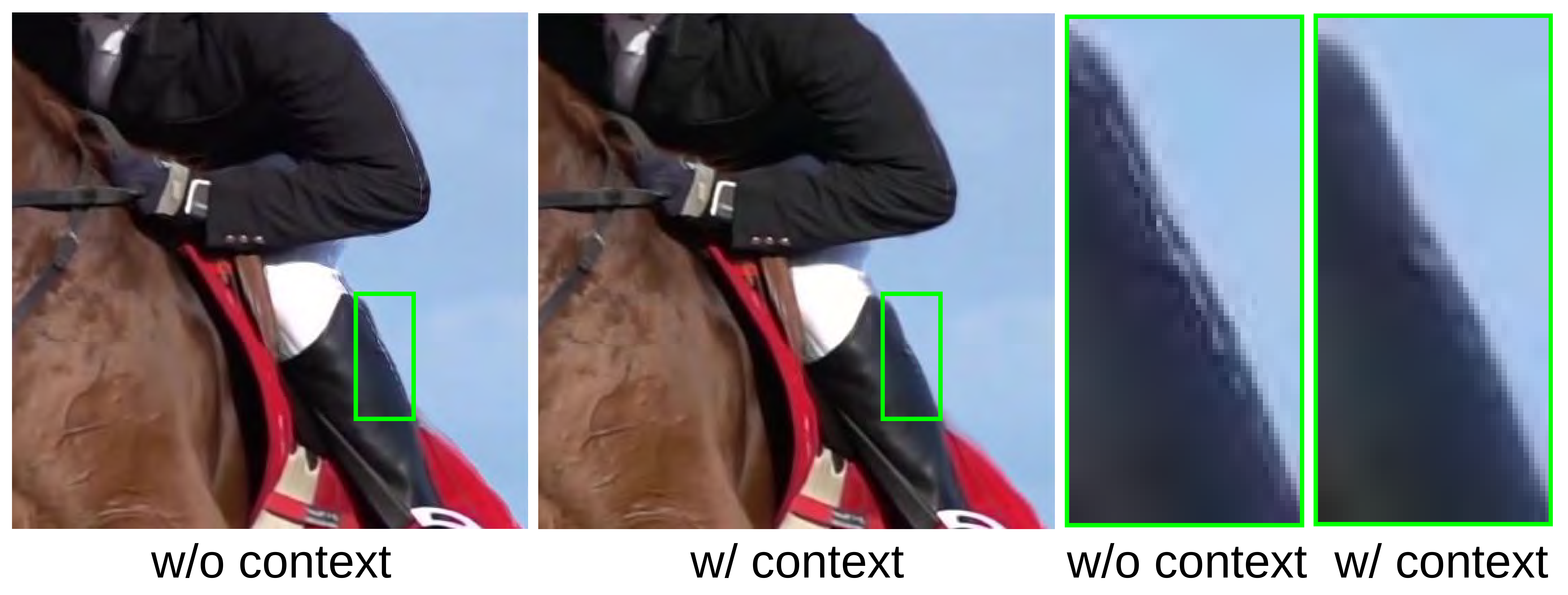}
 \vspace{-0.3in}
 \caption{Our appearance estimation without contextual information produces results with sharp artifacts at the motion boundaries (left). Using contextual information, we are able to reduce these artifacts (right).}
 \label{fig:AppearanceEffect}
 \vspace{-0.2in}
\end{figure}

The goal of this component is to generate the final frame from the warped keyframes and the target auxiliary frame. A straightforward approach is to use the warped keyframes and the target auxiliary frame as the input to the network to produce the final intermediate frame. However, this approach produces results with artifacts in the occluded regions, as shown in Fig.~\ref{fig:VisibilityEffect}. This is because the network has difficulty detecting the occluded regions from only the input warped frames. To address this issue, we mask the warped keyframes in the occluded regions by multiplying them with their corresponding visibility mask in a pixel-wise manner, i.e., instead of $g(I_r, F_r)$ and $g(I_l, F_l)$ we pass $V_l \odot g(I_l, F_l)$ and $V_r \odot g(I_r, F_r)$ as the input to the network.

Although incorporating visibility maps significantly reduces the artifacts in the occluded regions, the results still suffer from high frequency warping artifacts around the motion boundaries, as shown in Fig.~\ref{fig:AppearanceEffect}. These artifacts are caused by error in flow computation and error accumulation in chaining of low resolution flows. To address this issue, we add contextual information to our inputs to supply more information about the objects in the scene (like edges and object boundaries). Specifically, we extract contextual information from the keyframes and warp them along with the keyframes to use as input to our CNN. We use the feature maps from the \texttt{conv1} layer of ResNet-18~\cite{he2016deep, Niklaus18} (64 channels) to produce the contextual information. The two warped keyframes along with their warped contextual maps, produce a 134 channel input to the network. We also feed the upsampled auxiliary target frame and its contextual map (67 channels) as input to the network.

In summary, our appearance estimation network takes a 201 channel input and estimates the final high-resolution target frame (3 channels). We use the same encoder decoder architecture as flow enhancement network (Fig.~\ref{fig:Arch}), but with different inputs and output.

Note that, recent state-of-the-art video frame interpolation methods have either used visibility maps~\cite{Jiang18} or contextual information~\cite{DAIN,Niklaus18} to enhance the interpolation quality. In contrast, we propose to incorporate both of them to address the two key issues of occlusion and warping artifacts when combining the auxiliary and main frames.

%% file: Training.tex
\section{Training}
\label{sec:Implementation}

\subsection{Data}
\label{sec:Data}

We collect a set of 1080p (565 clips) and 720p (310 clips) videos at 240 fps from YouTube and use them along with the videos from Adobe240-fps dataset (118 clips)~\cite{Su17} as our training data. The videos have a variety of scenes with diverse objects (humans, vehicles, different terrains) and varying degrees of camera motion. From these videos, we extract a set of 23,000 shorter clips, each consisting of 12 frames. From each video clip, we randomly select 9 consecutive frames, use the first and last ones as the two main keyframes, and select the remaining 7 intermediate frames as ground truth. We then downsample the 9 frames by a factor of six and four for the 1080p and 720p videos, respectively, and use them as auxiliary frames with low resolution. To augment the data, we randomly change the direction of the sequence. We also randomly horizontally flip them and randomly crop each frame to patches of size $768\times 384$. Auxiliary patches (180p) are smaller by a factor of six or four depending on the keyframe resolution; we choose this factor to generalize our model to most real cameras.

\begin{figure}
 \includegraphics[width=\linewidth]{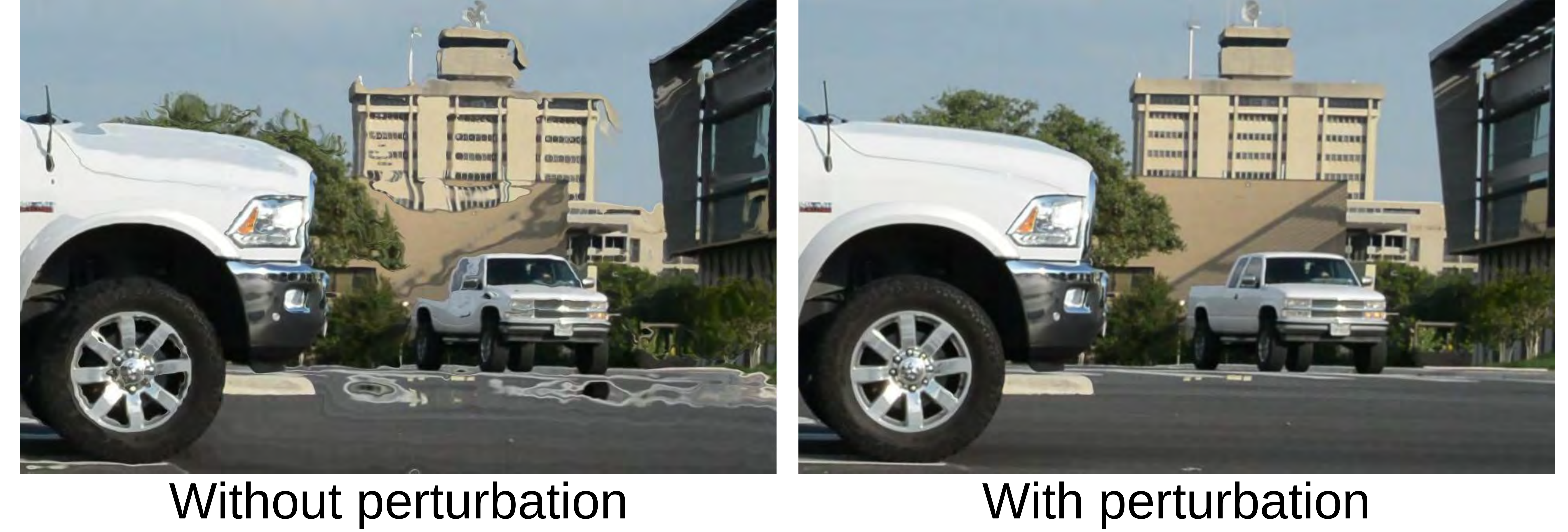}
 \vspace{-0.3in}
 \caption{Our system trained on synthetic data produces results with severe artifacts on the real data (left). By perturbing the synthetic training data, we simulate the imperfections of the real camera setup and are able to generate high-quality results on real data (right). }
 \label{fig:PerturbationEffect}
 \vspace{-0.15in}
\end{figure}

Our system trained on this synthetically generated dataset is not able to produce satisfactory results on videos captured with real hybrid \highlighttext{imaging} systems (see Fig.~\ref{fig:Rig}). This is mainly because the auxiliary frames in the training data are generated by directly downsampling the main video frames. However, in real dual camera setups, the auxiliary and main frames may have different brightness and color, and are captured from different viewpoints. To reduce the color and brightness variations, we use HaCohen et al.'s approach~\cite{NRDC2011} to match the color of the auxiliary frames to the main ones. While color transfer is generally helpful, it does not completely address the problem, as shown in Fig.~\ref{fig:PerturbationEffect}. Therefore, we propose a series of perturbations to increase the robustness of our system to brightness, color and viewpoint differences. First, we apply gamma correction with random $\gamma$ in the range [0.8, 1.3] to the auxiliary frames, before feeding them to our system. To increase the robustness of our approach to small misalignment between the main and auxiliary frames, we shift the auxiliary frames randomly from 0 to 2 pixels along both x and y axes before upsampling. As shown in Fig.~\ref{fig:PerturbationEffect}, these perturbations significantly improve the quality of the results on real data.

\vspace{-0.1in}
\subsection{Optimization}

We utilize four types of losses to train the flow enhancement and appearance estimation networks to improve quality of outputs at each stage and the reconstructed intermediate frame $\tilde{I}_t$. The pixel intensities of images lie in range [0, 1].
\vspace{0.05in}

\noindent\textbf{Reconstruction Loss $\mathcal{L}_r$}\quad One of the popular losses in image reconstruction is a pixel-wise color loss. We use the widely used pixel-wise $\mathcal{L}_1$ loss between the generated frame and ground truth frame~\cite{Niklaus17_CVPR, Niklaus18, Jiang18, Meyer18}.
\begin{equation}
    \mathcal{L}_r = \Vert \tilde{I}_t - I_t \Vert_1.
\end{equation}

\noindent\textbf{Perceptual Loss $\mathcal{L}_p$}\quad This loss helps generate fine details and improves the perceptual quality of the output image \cite{mathieu2015deep, Johnson2016Perceptual, zhang2018perceptual}. Specifically, we define the loss function as:
\begin{equation}
    \mathcal{L}_p = \Vert \phi(\tilde{I}_t) - \phi(I_t) \Vert_2^2,
\end{equation}
\noindent where $\phi$ is the response of \texttt{conv4\_3} layer of the pre-trained VGG-16 network~\cite{Simonyan15}, as commonly used in previous approaches ~\cite{Niklaus17_ICCV, Jiang18, Niklaus18}.
\vspace{0.1in}

\noindent\textbf{Warping Loss $\mathcal{L}_w$}\quad To improve the flows estimated by the flow enhancement network, we use a warping loss between the warped key frames and ground truth as:
\begin{equation}
    \mathcal{L}_w =\Vert I_t -  g(I_l, F_l) \Vert_1 + \Vert I_t - g(I_r, F_r) \Vert_1.
\end{equation}
\noindent\textbf{Total variation loss $\mathcal{L}_{tv}$}\quad To enforce the estimated flow to be smooth\cite{Liu17, Jiang18}, we also minimize the total variation loss on the estimated flows as:
\begin{equation}
    \mathcal{L}_{tv} =\Vert \nabla F_l \Vert_1 + \Vert \nabla F_r \Vert_1.
\end{equation}
Training the flow enhancement and appearance estimation networks by directly minimizing the error between the estimated and ground truth final frames is difficult. Therefore, we propose to perform the training in three stages. In the first stage, we only train the flow enhancement network without the need for the ground truth flows and visibility maps \highlighttext{by minimizing the following loss function:}
\begin{equation}
    \mathcal{L}_{align} = 204 \ \mathcal{L}_r + 0.005 \ \mathcal{L}_p + 102 \ \mathcal{L}_{w} + \mathcal{L}_{tv}.
\end{equation}

\noindent \highlighttext{Note that, $\mathcal{L}_r$ computes the distance between the estimated and ground truth frames. We compute the target frame using the estimated flows and visibility maps as follows:}

\begin{equation}
\tilde{I}_t = \frac{(1 - t) V_l \ g(I_l, F_l) + t V_r \ g(I_r, F_r)}{(1 - t) V_l + t V_r}.
\end{equation}

\highlighttext{In this case, the flow network is trained to estimate flows and visibility maps that result in high-quality images.} Once the network is converged, in the second stage, we only train the appearance estimation network by minimizing the appearance loss between the estimated and ground truth target frames as given below,
\begin{equation}
    \mathcal{L}_{appearance} = 204 \ \mathcal{L}_r + 0.005 \ \mathcal{L}_p.
\end{equation}
Note that, the estimated target frame in this stage is the output of the appearance estimation network. At the end, we fine-tune both the networks jointly using only the perceptual loss $\mathcal{L}_p$ to improve the details in $\tilde{I}_t$.

%% file: Results.tex
\section{Results}

We implemented our model using PyTorch and used the ADAM~\cite{Kingma14} optimizer to train the networks. We train the flow enhancement network for 300 epochs, using an initial learning rate of $10^{-4}$ and decreased the rate by a factor of 10 every 100 epochs. Similarly, we train the appearance estimation network for 75 epochs, with an initial learning rate of $10^{-5}$ and decreased the rate by a factor of 10 every 25 epochs. 

\begin{table}[!t]
\renewcommand{\arraystretch}{1.3}
\caption{Results on Slow Flow~\cite{Janai2017CVPR}, Middlebury~\cite{Baker11} and NfS~\cite{galoogahi2017need} datasets}
\vspace{-0.15in}
\centering
\begin{tabular}{lcccccc}
  \hline\hline
  & \multicolumn{2}{c}{Slow Flow} & \multicolumn{2}{c}{Middlebury} & \multicolumn{2}{c}{NfS}\\
  \cmidrule(lr){2-3}\cmidrule(lr){4-5}\cmidrule(lr){6-7}
  & LPIPS & SSIM & LPIPS & SSIM & LPIPS & SSIM\\
  \hline
   DUF & 0.4076 & 0.556 & 0.3410 & 0.585 & 0.2215 & 0.745\\
   EDVR & 0.3860 & 0.657 & 0.2882 & 0.670 & 0.1751 & 0.839\\
   DAIN & 0.1585 & 0.832 & 0.1709 & 0.729 & 0.0626 & 0.928\\ 
   Super SloMo & 0.1525 & 0.833 & 0.1498 & 0.736 & 0.0563 & 0.929\\
   \hline
   Ours & \textbf{0.1332} & \textbf{0.865} & \textbf{0.1107} & \textbf{0.807} & \textbf{0.0467} & \textbf{0.956}\\
  \hline\hline
\end{tabular}
\label{tab:Results}
\vspace{-0.2in}
\end{table}

We compare our approach against state-of-the-art multi-frame video interpolation and video super-resolution deep learning methods on a set of synthetic and real videos. Super SloMo~\cite{Jiang18} and DAIN~\cite{DAIN} increase the frame rate of the main video, and EDVR~\cite{wang2019edvr} and DUF~\cite{Jo_2018_CVPR} increase the spatial resolution of the auxiliary video. We also compare against Gupta et al.'s~\cite{gupta:2009} non-learning approach, which similar to our system, utilizes both the main and auxiliary frames to generate the output frames. Since the source code for Super SloMo is not available, we implemented it ourselves. We replaced the \texttt{flow computation} network with PWC-Net~\cite{Sun18} as it gives better performance, especially on videos with high resolution and large motion. For Gupta et al.'s approach, we use one of the scenes from their supplementary video for comparison. For all the other approaches, we use the source code provided by the authors.

\begin{figure}
  \includegraphics[width=\linewidth]{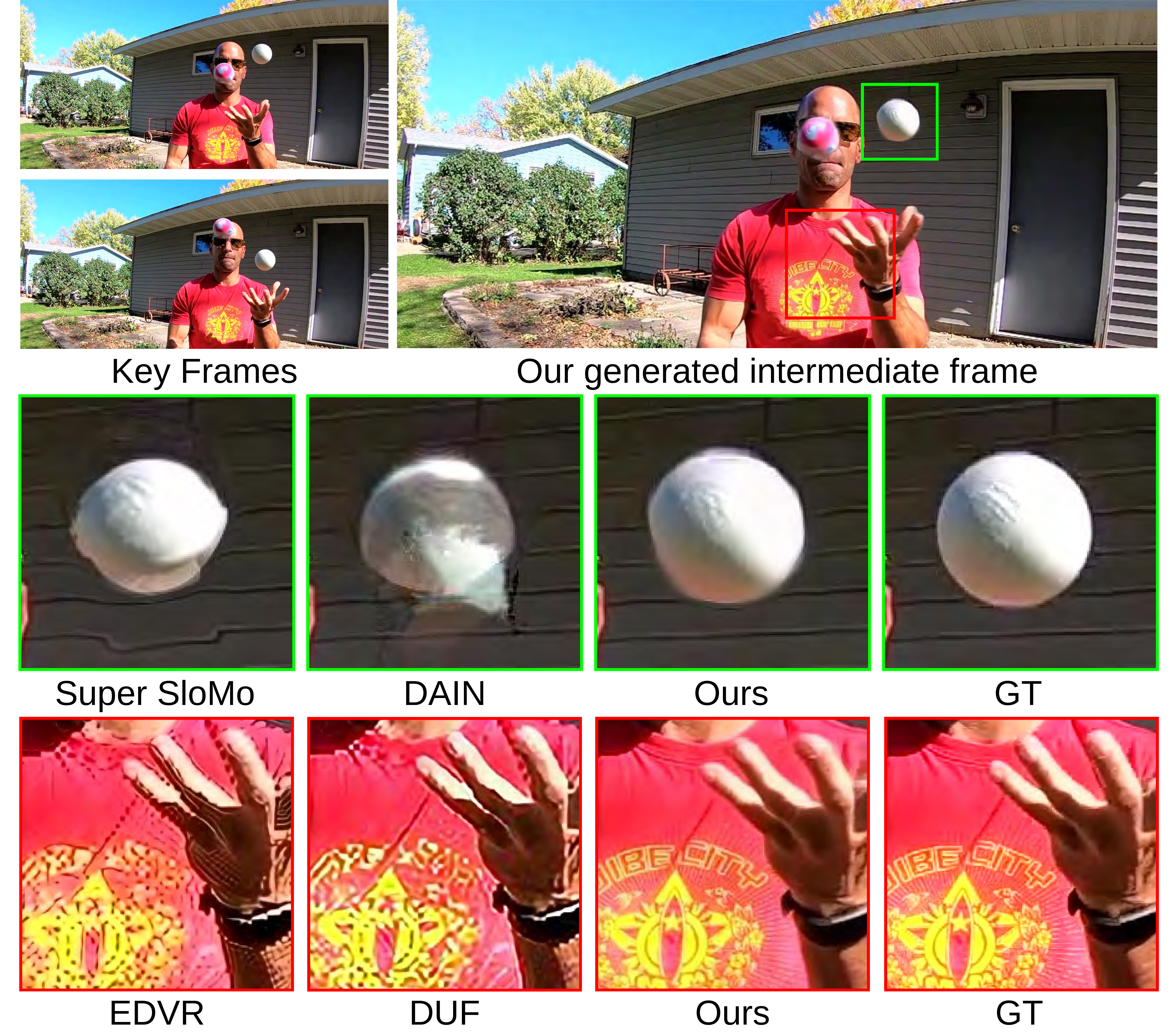}
  \includegraphics[width=\linewidth]{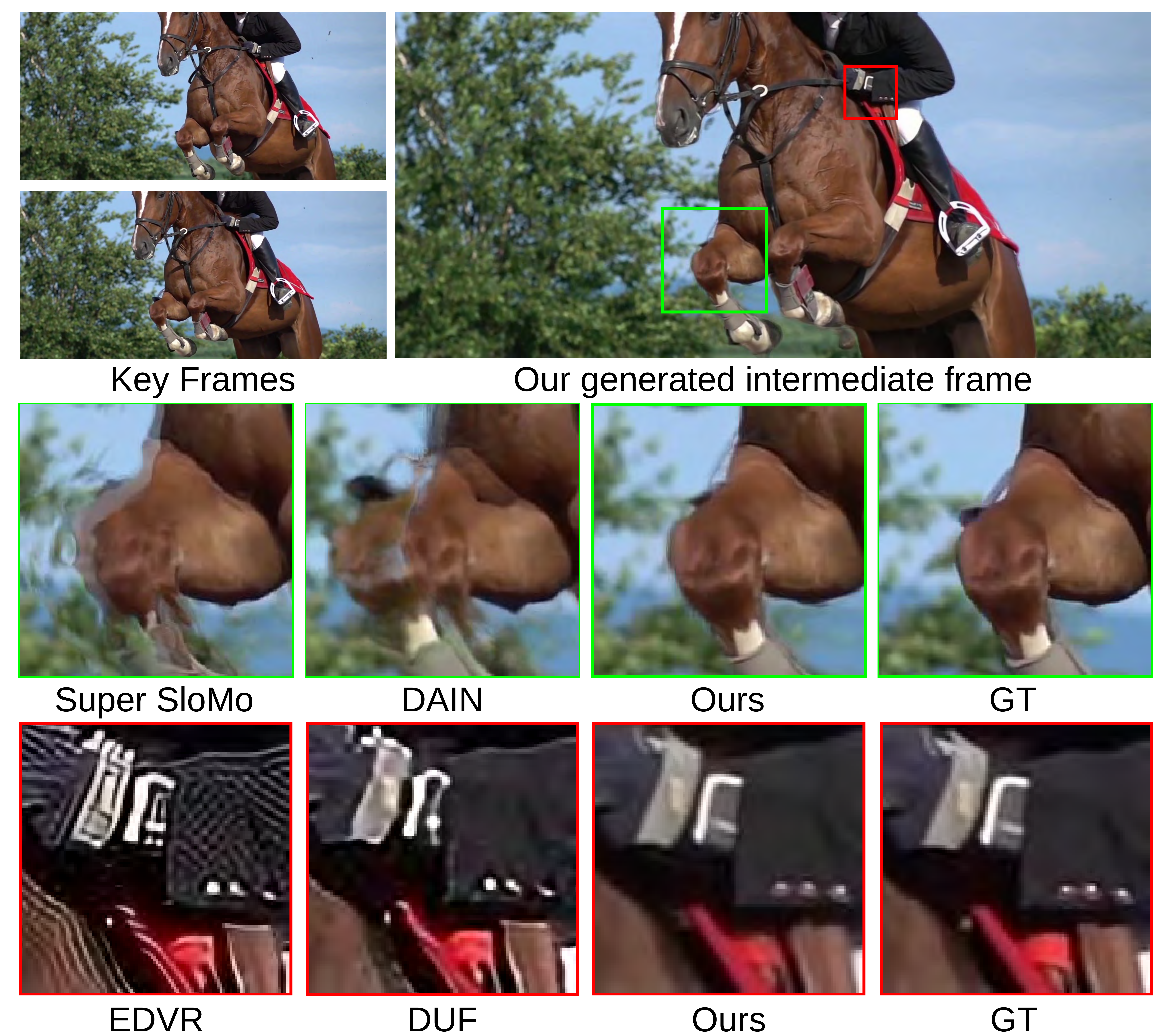}
  \vspace{-0.3in}
  \caption{Comparison against state-of-the-art multi frame video interpolation and video super-resolution methods on \textsc{Juggler} and \textsc{Horse} scenes. The auxiliary video in this case has been synthetically generated from the main video.}
  \label{fig:SynResults}
  \vspace{-0.2in}
\end{figure}

\subsection{Synthetic Videos} We begin by quantitatively evaluating our results on the Slow Flow~\cite{Janai2017CVPR}, Middlebury~\cite{Baker11} and (NfS)~\cite{galoogahi2017need} datasets. We use the 20 color sequences from the \texttt{eval-color} and \texttt{other-color} Middlebury datasets, each with 8 frames. We use the first and last frames as keyframes and interpolate the 6 frames in between. For the 41 sequences in the Slow Flow dataset and 9 test sequences from NfS dataset, we interpolate 7 in-between frames. To generate the auxiliary frames, we downsample the original frames by a factor of 3 along each dimension. As shown in Table.~\ref{tab:Results}, our method produces better results than the state-of-the-art methods in terms of two perceptual metrics, SSIM~\cite{ssim} and LPIPS~\cite{zhang2018perceptual}.

In Fig.~\ref{fig:SynResults}, we compare our approach with state-of-the-art methods on two 1080p test videos. These are 240 fps videos and we attempt to interpolate 12 frames, effectively increasing the frame rate from 20 to 240 fps. We generate the auxiliary frames by downsampling the input frames by a factor of 4 in each dimension. The \textsc{Juggler} scene (top) contains large motion on the ball, as seen in the keyframes. The video interpolation methods of DAIN and Super SloMo are not able to properly handle this challenging scene, producing artifacts on the ball. The video super-resolution methods of EDVR and DUF are unable to recover the texture details on the shirt. Our approach correctly handles regions with motion and is able to recover high resolution texture.

The \textsc{Horse} scene (bottom) is challenging because of the large motion of the horse and relative motion of objects on the horse (reins, rider). DAIN and Super SloMo generate results with severe artifacts in the fast moving areas. Since most of the background is out of focus and the foreground has smooth texture (horse, saddle), EDVR and DUF are able to produce reasonable results in these areas. However, they are unable to completely recover the intricate details on rider's clothes and produce results with artifacts on thin objects like reins. By utilizing both the main and auxiliary frames, our method is able to reconstruct results with fine details and correct motions.

\begin{figure}
  \includegraphics[width=\linewidth]{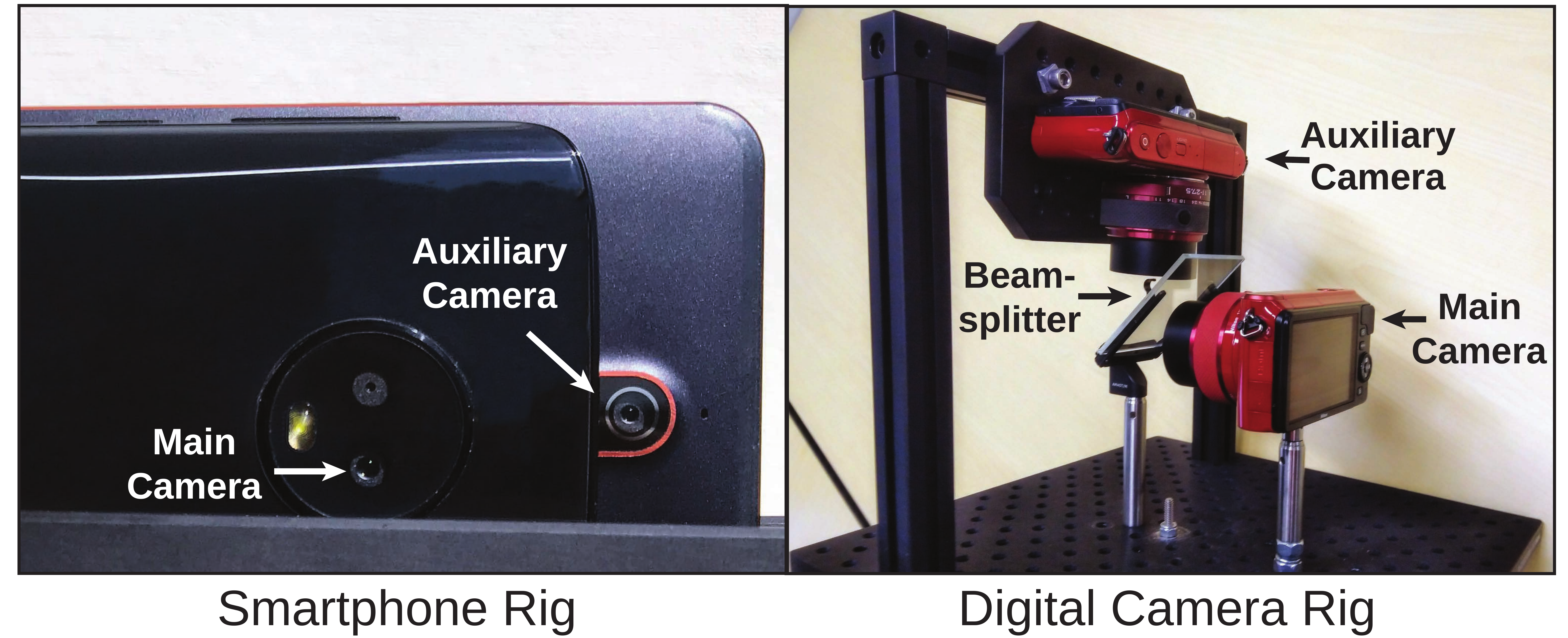}
  \vspace{-0.3in}
  \caption{Prototypes of our hybrid \highlighttext{imaging} system. a) The smartphone rig (left) is a simple setup with Moto G6 (main camera) and Nokia 6.1 (auxiliary camera). b) The digital camera rig (right) is designed using two Nikon S1 and a 50:50 beam-splitter to emulate a small baseline setup.}
  \label{fig:Rig}
  \vspace{-0.2in}
\end{figure}

\begin{figure}
  \includegraphics[width=\linewidth]{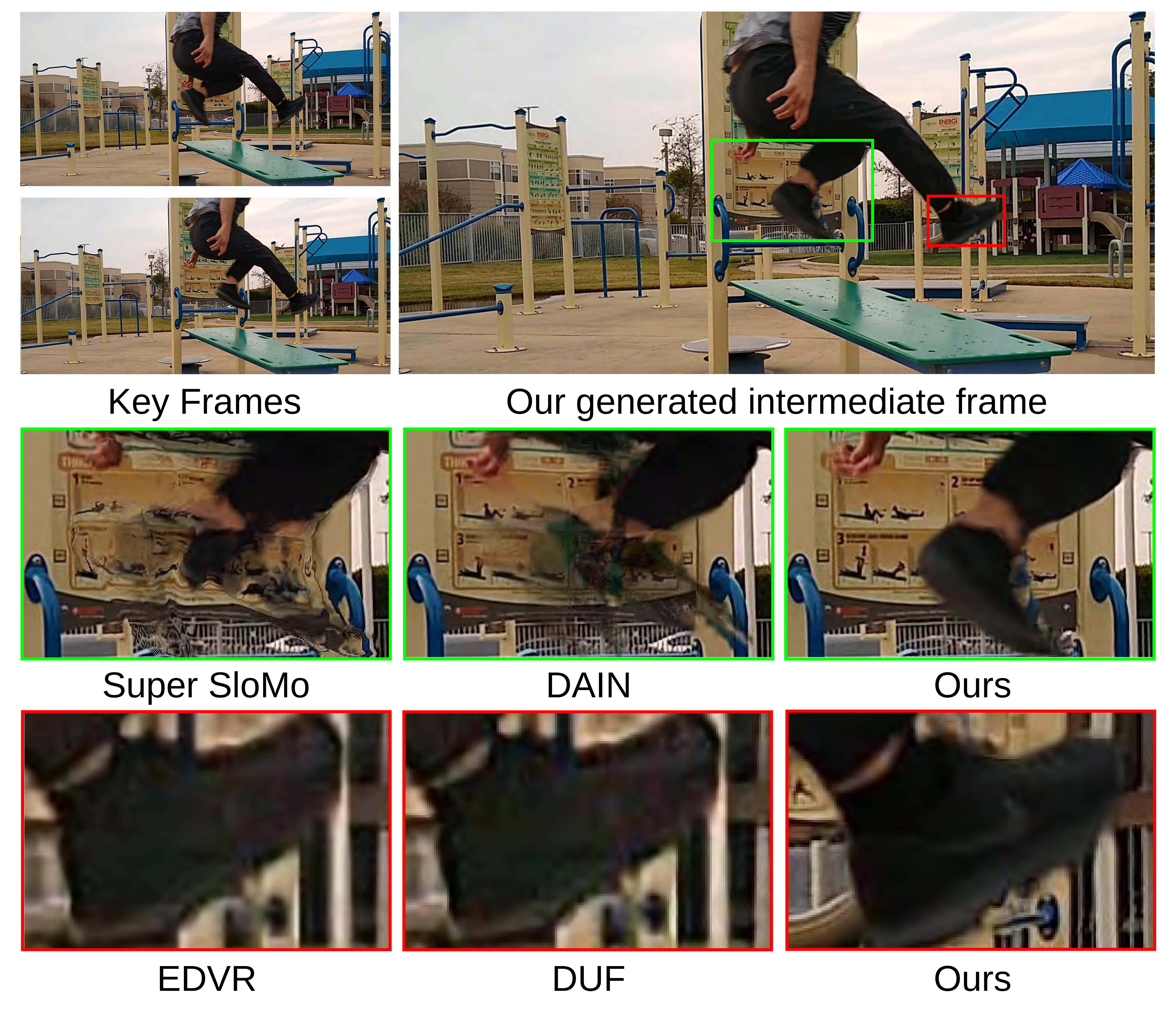}
  \vspace{0.05in}
  \includegraphics[width=\linewidth]{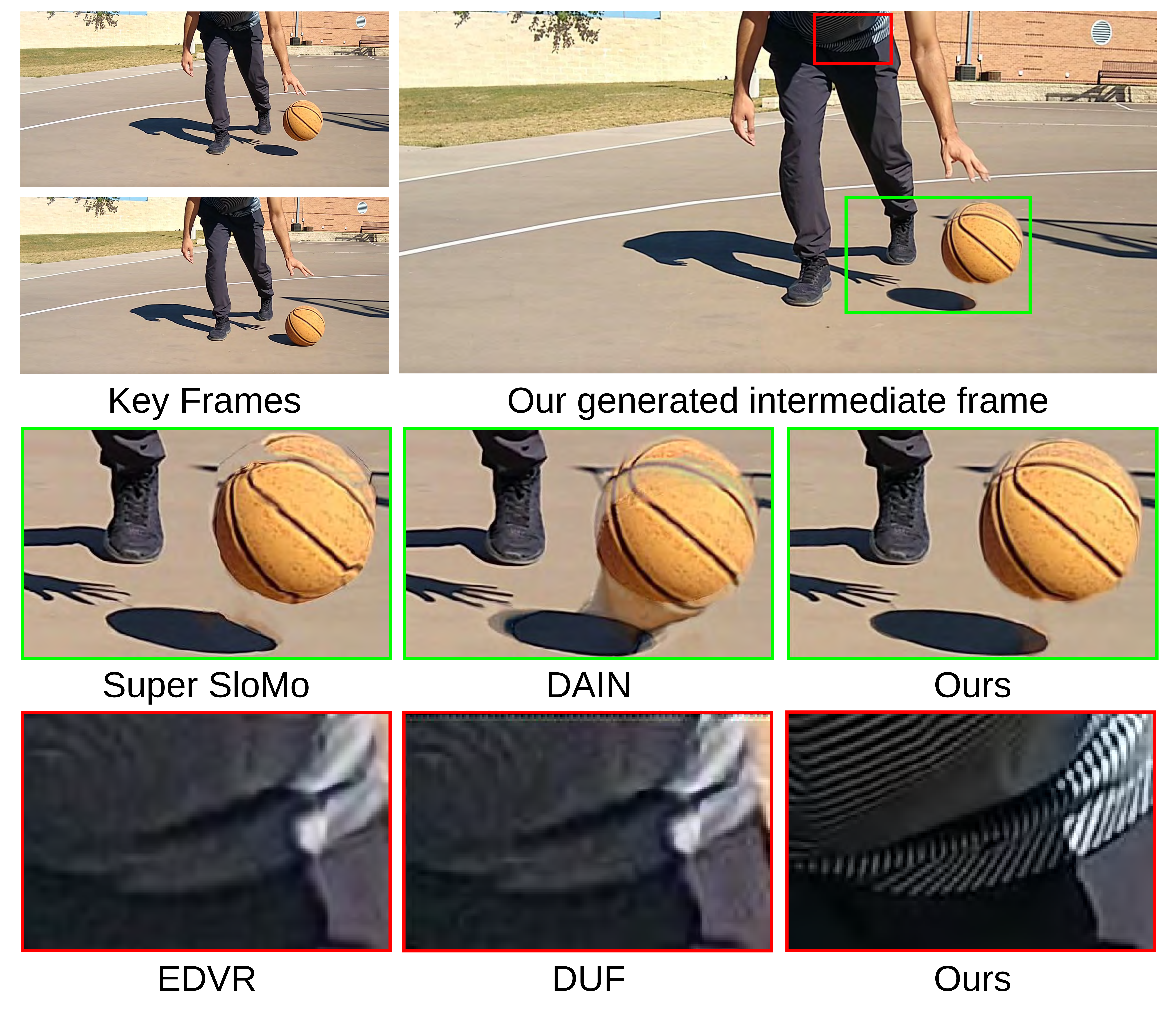}
  \includegraphics[width=\linewidth]{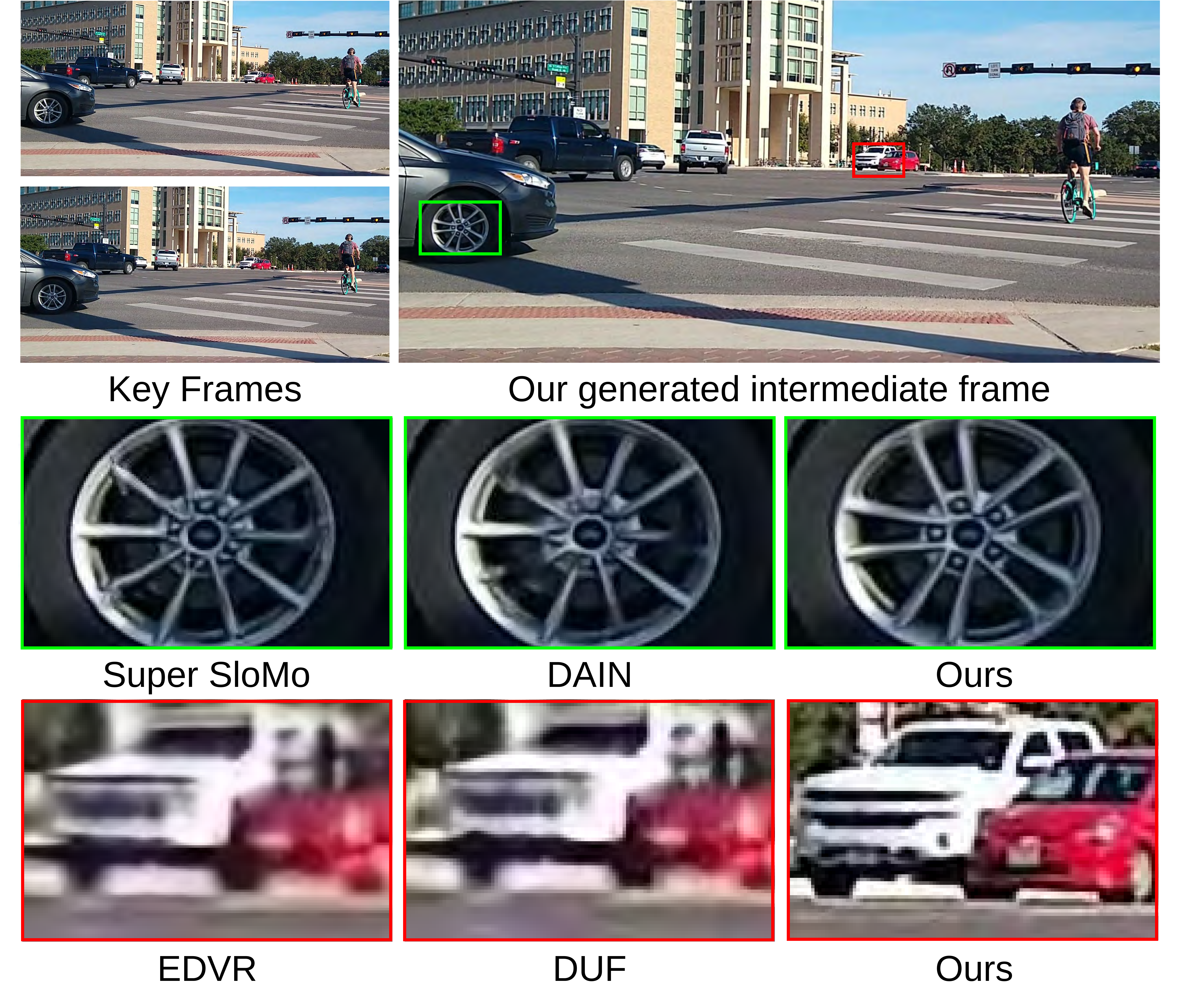}
  \vspace{-0.25in}
  \caption{Comparison against state-of-the-art multi frame video interpolation and video super-resolution methods on the \textsc{Jump} (top), \textsc{Dribble} (middle), and \textsc{Car} (bottom) sequences captured with our hybrid smartphone camera rig.}
  \label{fig:RealResultsSmartphone}
\end{figure}

\begin{figure}
  \includegraphics[width=\linewidth]{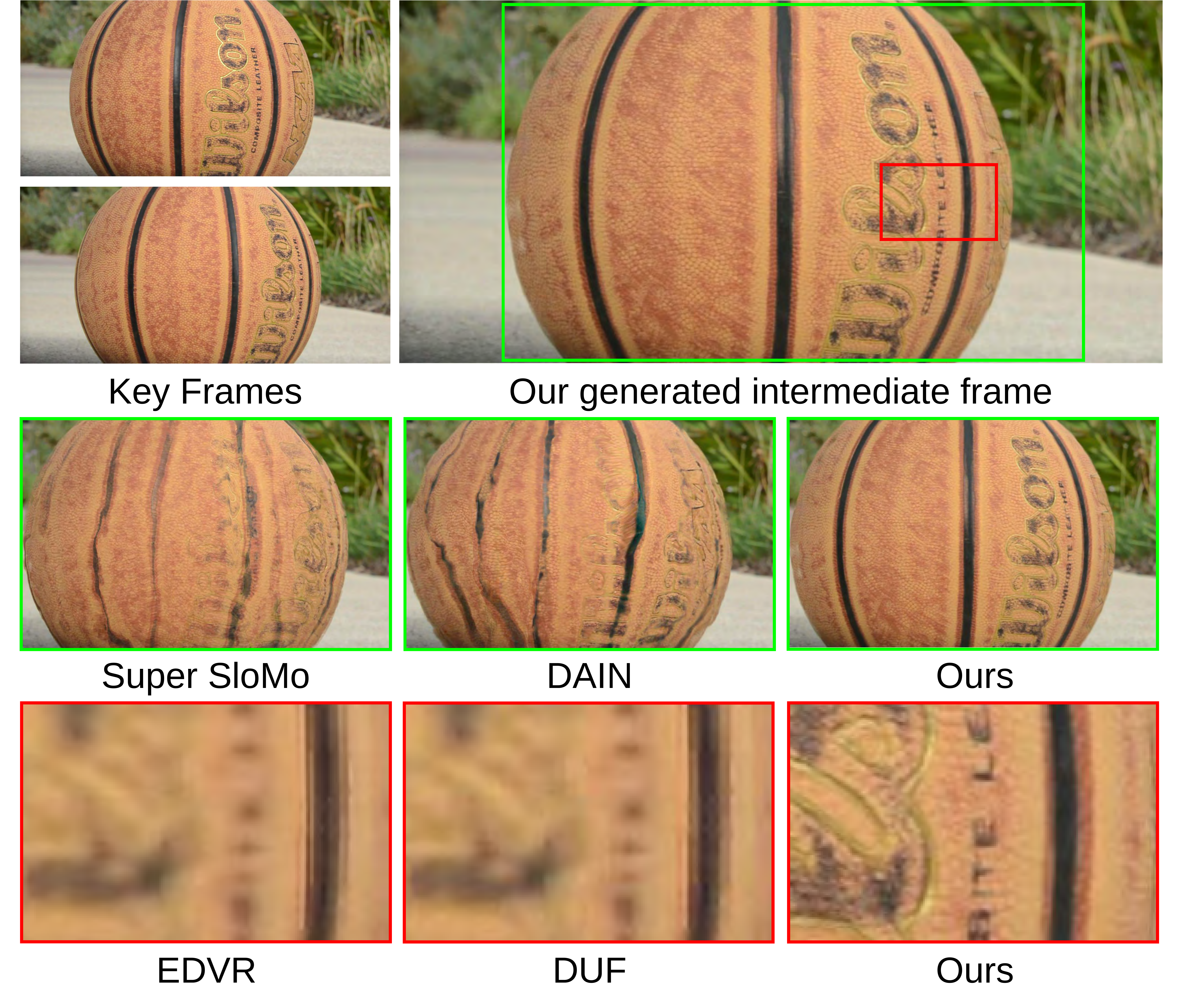}
  \includegraphics[width=\linewidth]{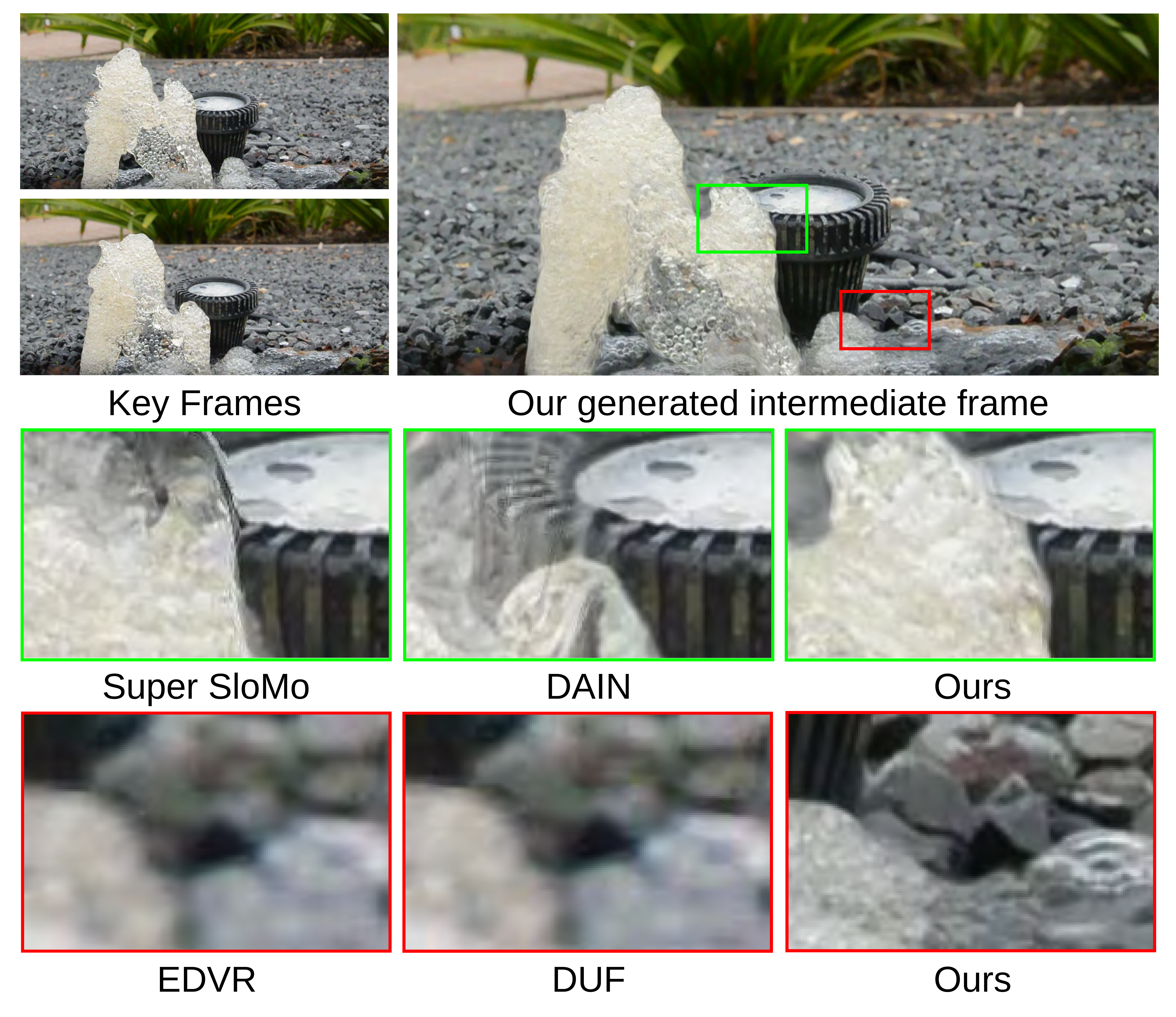}
  \includegraphics[width=\linewidth]{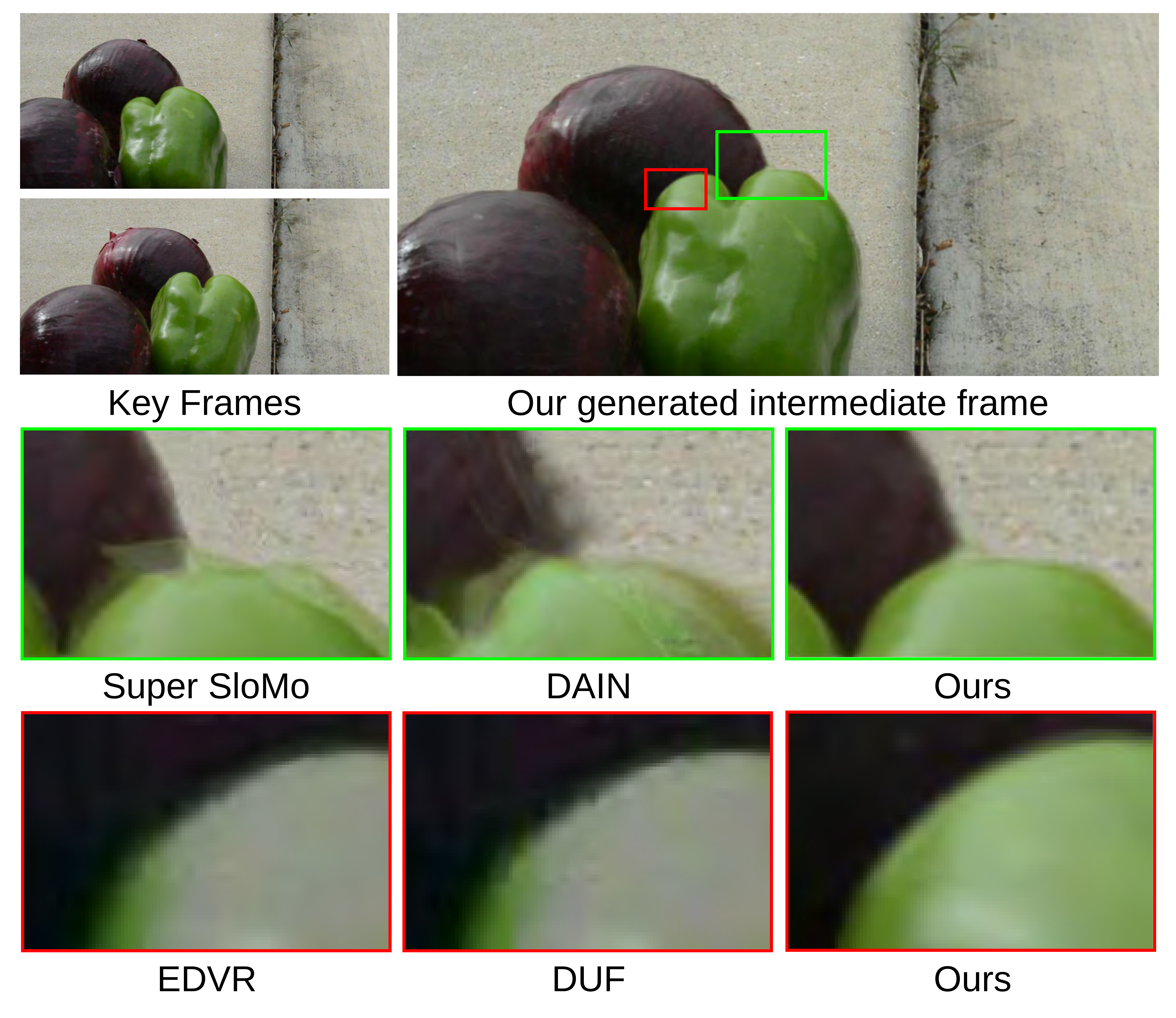}
  \vspace{-0.2in}
  \caption{Comparison against state-of-the-art multi frame video interpolation and video super-resolution methods on the \textsc{Basketball} (top), \textsc{Fountain} (middle), and \textsc{Veggies} (bottom) sequences captured with our hybrid digital camera rig.}
  \label{fig:RealResultsDigCam}
\end{figure}

\vspace{-0.1in}
\subsection{Real Camera Setups}\label{sec:RealCameras} To capture the real videos with different resolutions and frame rates, we construct two hybrid \highlighttext{imaging} systems, as shown in Fig.~\ref{fig:Rig}.
The first rig is simple with two low end smartphones, Moto G6 and Nokia 6.1, placed side-by-side. The main camera captures 30 fps videos with a resolution of $1920\times 1080$ and auxiliary camera records 120 fps videos of effective size $570\times 320$; we crop the original auxiliary video ($720\times 480$) to match the field of view of the main camera. \highlighttext{For all our real results, the cameras were manually triggered without any synchronization. During preprocessing, we simply discard the frames from one video to match the first frame of the other video. While this step reduces the mismatch between the main and auxiliary frames, small temporal misalignments remain between the corresponding frames. However, our system properly handles these cases since we train our flow network to be robust to small misalignments (see Sec. \ref{sec:Data}). Moreover, both our setups use cameras with rolling shutters and, thus, the captured videos have distortions on fast-moving objects. However, our flow enhancement network treats these distortions as misalignment between the auxiliary and main videos and estimates appropriate residual flows.}

From the corresponding results in Fig.~\ref{fig:RealResultsSmartphone}, Super SloMo and DAIN are not able to properly handle regions with \highlighttext{motion blur and} large motion like the leg and basketball in \textsc{Jump} and \textsc{Dribble} scenes. A particularly interesting case is the \textsc{Car} scene where Super SloMo and DAIN produce results with temporal aliasing on the wheel (see supplementary video). EDVR and DUF, on the other hand, are not able to recover the sharp details in the textured areas like the shirt in the \textsc{Dribble} scene. Our method is able to effectively utilize information from the main and auxiliary videos to generate high quality results. These results demonstrate that we can use the dual camera setups in most recent smartphones to capture high resolution slow motion videos.

\begin{figure}
  \includegraphics[width=\linewidth]{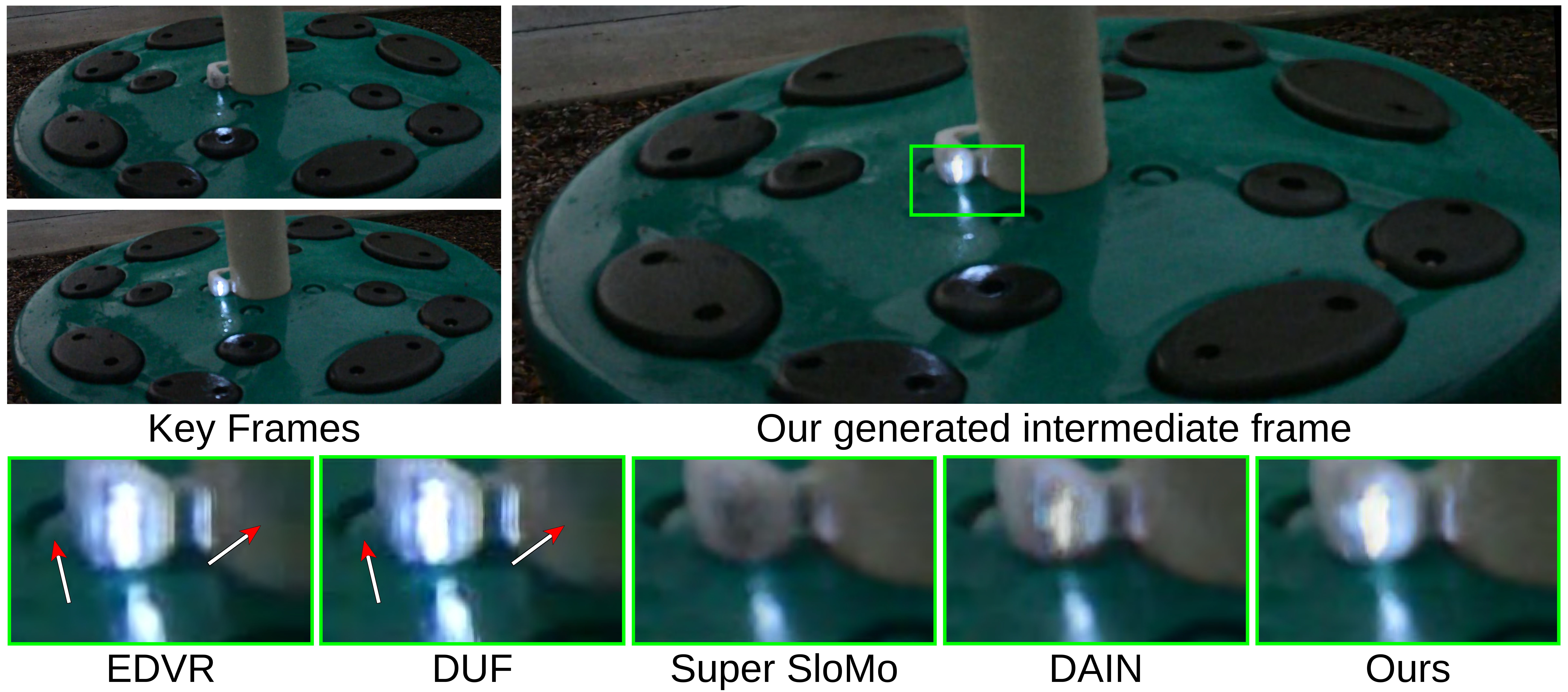}
  \vspace{-0.3in}
  \caption{\highlighttext{This scene is captured under low light condition and, thus, the auxiliary frames are noisy. EDVR and DUF are unable to reconstruct the details (left arrow) and remove the input noise (right arrow). Due to the rolling shutter effect, the light torch is off but the reflection is visible in the left keyframe, which is problematic for Super SloMo and DAIN. Our method, however, is able to handle this challenging scene.}}
  \label{fig:NoiseReal}
  \vspace{-0.25in}
\end{figure}

To demonstrate our approach's ability to generate videos with very high frame rates, we construct another rig with two Nikon S1 cameras. One captures 30 fps videos with resolution of $1920\times 720$ and the other records 400 fps videos of size $640\times 240$. To reduce the baseline of the cameras, we place a 50:50 beam-splitter at 45$^{\circ}$ angle between them.

The \textsc{Basketball} scene from Fig.~\ref{fig:RealResultsDigCam} shows a basketball rotating to the right. Super SloMo and DAIN are unable to capture the flow of this rotation and, therefore, produce results with artifacts. EVDR and DUF are not able to recover details like writing on the basketball. However, our method generates a high quality video since it can properly utilize the temporal information from the auxiliary frames. Next, we examine the \textsc{Fountain} scene, which has highly complex and stochastic motion. Super SloMo and DAIN produce a video with unnatural motion (see supplementary video). EVDR and DUF produce blurry results, lacking details  on the rocks and water. Our method generates a coherent video with natural motion and detailed texture. The \textsc{Veggies} scene is particularly challenging for our method since auxiliary video contains saturated areas. Super SloMo and DAIN are unable to capture the rotation of the vegetables and simply translate them horizontally. EDVR and DUF produce noisy results and are unable to recover the saturated areas in the auxiliary frames. Our method correctly captures the motion of the vegetables and recovers the saturated content by utilizing the main keyframes.

\highlighttext{We also demonstrate the performance of our method on a scene captured under low light condition in Fig.~\ref{fig:NoiseReal}. Note that, the auxiliary frames in this case are noisy. In addition to producing blurry results, EDVR and DUF are not able to remove the noise. This scene also demonstrate the rolling shutter effect where the light reflection appears before the torch turns on (see supplementary video). Super SloMo and DAIN are not able to handle this case and generate artifacts in these regions. Our method effectively utilizes the information in the auxiliary video to generate the output.
}

\vspace{-0.1in}
\subsection{Comparison against Gupta et al.}
\begin{figure}
  \includegraphics[width=\linewidth]{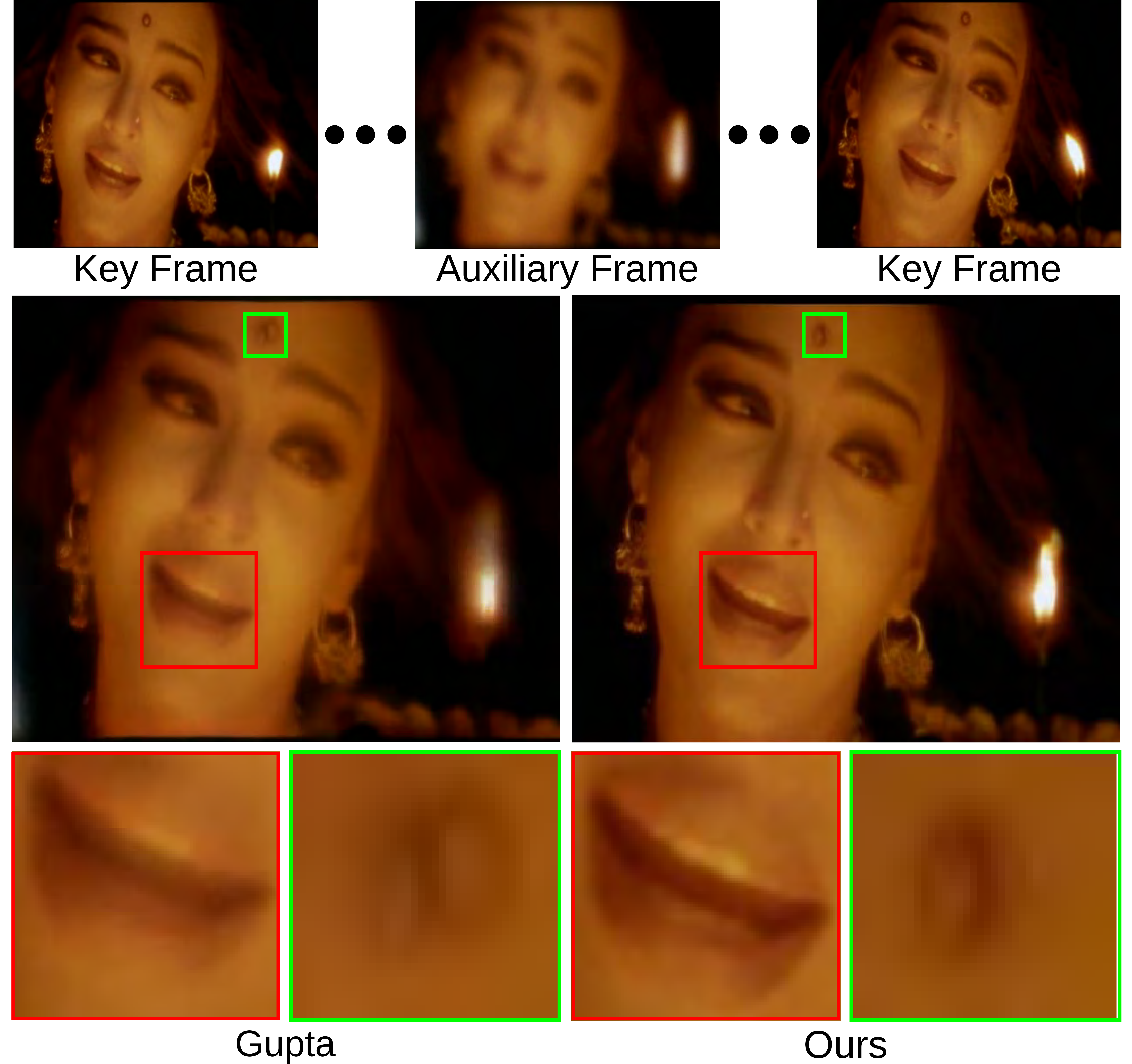}
  \vspace{-0.3in}
  \caption{Comparison against Gupta et al.'s~\cite{gupta:2009} approach on the \textsc{Lady} scene from their supplementary video. This is a challenging scene with constant change in lighting due to the flickering of lamp.}
  \label{fig:GuptaComp}
  \vspace{-0.15in}
\end{figure}

We compare our approach with the non-learning method by Gupta et al.~\cite{gupta:2009} on a synthetic scene (Fig.~\ref{fig:GuptaComp}). The input videos for their results are not available and, therefore, we extract the main and auxiliary videos from their supplementary video. The extracted \textsc{Lady} scene is challenging due to constant change in lighting caused by the flickering of lamp. Gupta et al.'s method has a trade-off between spatial resolution and temporal consistency and, thus, produces results with blurry patches. In comparison, our method produces a high quality output with minimal ghosting artifacts.

\begin{table}[!t]
\renewcommand{\arraystretch}{1.3}
\caption{\highlighttext{Inference performance on GTX 1080 Ti GPU}}
\vspace{-0.15in}
\centering
\begin{tabular}{l | c c c c c}
  \hline\hline
   \# Intermediate Frames & 9 & 7 & 5 & 3 & 1\\
   \hline
   Run-time (seconds/frame) & 0.90 & 0.87 & 0.84 & 0.70 & 0.22\\
  \hline\hline
\end{tabular}
\label{tab:InferencePerformance}
\vspace{-0.15in}
\end{table}

\begin{table}[!t]
\renewcommand{\arraystretch}{1.3}
\caption{Evaluating the effect of visibility and contextual information in appearance estimation}
\vspace{-0.15in}
\centering
\begin{tabular}{lcc}
  \hline\hline
  \multicolumn{1}{c}{Component} &  LPIPS & SSIM\\
  \hline
   Appearance Estimation & 0.1193 & 0.891\\
    \quad+ Visibility Maps & 0.1110 & 0.896\\
   \quad+ Contextual Information & \textbf{0.1012} & \textbf{0.903}\\
  \hline\hline
\end{tabular}
\label{tab:AblationComponent}
\vspace{-0.2in}
\end{table}

\vspace{-0.1in}
\subsection{Inference performance}
\highlighttext{ Table~\ref{tab:InferencePerformance} shows the inference performance of our model on a GTX 1080 Ti GPU using a main and auxiliary video sequence of resolutions 1080p and 360p, respectively. Note that, the average time for generating an intermediate frame increases with the number of intermediate frames to be interpolated. This is primarily caused by the chaining of flows between consecutive auxiliary frames during the initial flow estimation (Sec.~\ref{sec:InitFlow}).}

\vspace{-0.1in}
\subsection{\highlighttext{Analysis}}
\subsubsection{Effect of Visibility and Contextual Information}
We study the impact of different components in appearance estimation. Specifically, we evaluate the effect of \textit{visibility maps} and \textit{contextual information} on the quality of the output of appearance estimation network. Here, we use 30 fps 1080p main and 240 fps 270p auxiliary videos. As shown in Table~\ref{tab:AblationComponent}, visibility maps improve the result as indicated by LPIPS and SSIM. Adding contextual information further enhances the output by providing perceptual and semantics information about the scene (edges, object boundaries).

\begin{table}[!t]
\renewcommand{\arraystretch}{1.3}
\caption{Evaluating the effect of auxiliary video resolution}
\vspace{-0.15in}
\centering
\begin{tabular}{l c c c c c c}
  \hline\hline
  & 540p & 360p & 270p & 180p & 90p & 45p \\ 
  \hline
   LPIPS & \textbf{0.1005} & \textbf{0.1005} & 0.1012 & 0.1065 & 0.1365 & 0.2011\\
   SSIM & \textbf{0.906} & \textbf{0.906} & 0.903 & 0.892 & 0.851 & 0.811\\
  \hline\hline
\end{tabular}
\label{tab:AblationPSNR}
\vspace{-0.15in}
\end{table}

\begin{table}[!t]
\renewcommand{\arraystretch}{1.3}
\caption{Evaluating the effect of main video frame rate}
\vspace{-0.15in}
\centering
\begin{tabular}{l c c c c c}
  \hline\hline
  & 120 fps & 60 fps & 40 fps & 30 fps & 24 fps \\ 
  \hline
   LPIPS & \textbf{0.0535} & 0.0709 & 0.0856 & 0.0976 & 0.1095\\
   SSIM & \textbf{0.951} & 0.936 & 0.923 & 0.911 & 0.901\\
  \hline\hline
\end{tabular}
\label{tab:AblationKey}
\vspace{-0.15in}
\end{table}

\vspace{-0.1in}
\subsubsection{\highlighttext{Auxiliary Video Resolution}}
Here, we evaluate the performance of our model with varying auxiliary video resolutions. To do this, we use main and auxiliary videos at 30 and 240 fps, respectively. We keep the resolution of the main video at 1080p, and evaluate the quality of reconstructed video by changing the resolution of the auxiliary video from 45p to 540p. The results in Table~\ref{tab:AblationPSNR} show that the performance degrades sharply below 180p and improvement above it is minimal. This shows that our model can effectively extract temporal information from a very low auxiliary frame resolution of 180p.

\vspace{-0.1in}
\subsubsection{\highlighttext{Main Video Frame Rate}}
In this section, we analyze the effect of main video (1080p) frame rate on the output of the network. Here, we use auxiliary videos with 240 fps and resolution of 360p. We vary the main video's frame rate from 24 fps to 120 fps and interpolate one middle frame (same middle frame for all experiments). As shown in Table~\ref{tab:AblationKey}, the network performs better by increasing the frame rate of the main video, as expected. Higher frame rate effectively reduces the displacement of objects in consecutive keyframes which results in higher quality flows and output.

\begin{table}[!t]
\renewcommand{\arraystretch}{1.3}
\caption{\highlighttext{Evaluating output quality by changing gamma and hue of the auxiliary video frames}}
\vspace{-0.15in}
\centering
\begin{tabular}{l c c c c c c}
  \hline\hline
  Gamma & 0.65 & 0.85 & 0.95 & 1.00 & 1.25 & 1.75\\ 
  \hline
   LPIPS & 0.1016 & \textbf{0.1005} & \textbf{0.1005} & \textbf{0.1005} & 0.1006 & 0.1052\\
   SSIM & 0.905 & \textbf{0.906} & \textbf{0.906} & \textbf{0.906} & 0.905 & 0.903\\
   \hline
   Hue & 0.00 & 0.02 & 0.05 & 0.15 & 0.30 & 0.50\\ 
   \hline
   LPIPS & \textbf{0.1005} & 0.1005 & 0.1007 & 0.1014 & 0.1018 & 0.1018\\
   SSIM & \textbf{0.906} & 0.906 & 0.905 & 0.905 & 0.905 & 0.904\\
  \hline\hline
\end{tabular}
\label{tab:AblationTMP}
\vspace{-0.15in}
\end{table}

\vspace{-0.1in}
\subsubsection{\highlighttext{Differences in Color and Brightness}}
\label{sec:gammahue}
\highlighttext{Here, we evaluate the robustness of our model to differences in brightness and color between main and auxiliary videos. To do this, we evaluate the quality of the reconstructed frames by changing gamma and hue of the auxiliary videos (see Fig.~\ref{fig:ExpInp}). Table~\ref{tab:AblationTMP} shows the result of this experiment on main (1080p) and auxiliary (360p) videos with frame rates of 30 and 240. As seen, our method performs consistently well for a large range of gamma and hue variations.}


\vspace{-0.1in}
\subsubsection{\highlighttext{Effect of Noise}}
\label{sec:noisy}
\begin{figure}
  \includegraphics[width=\linewidth]{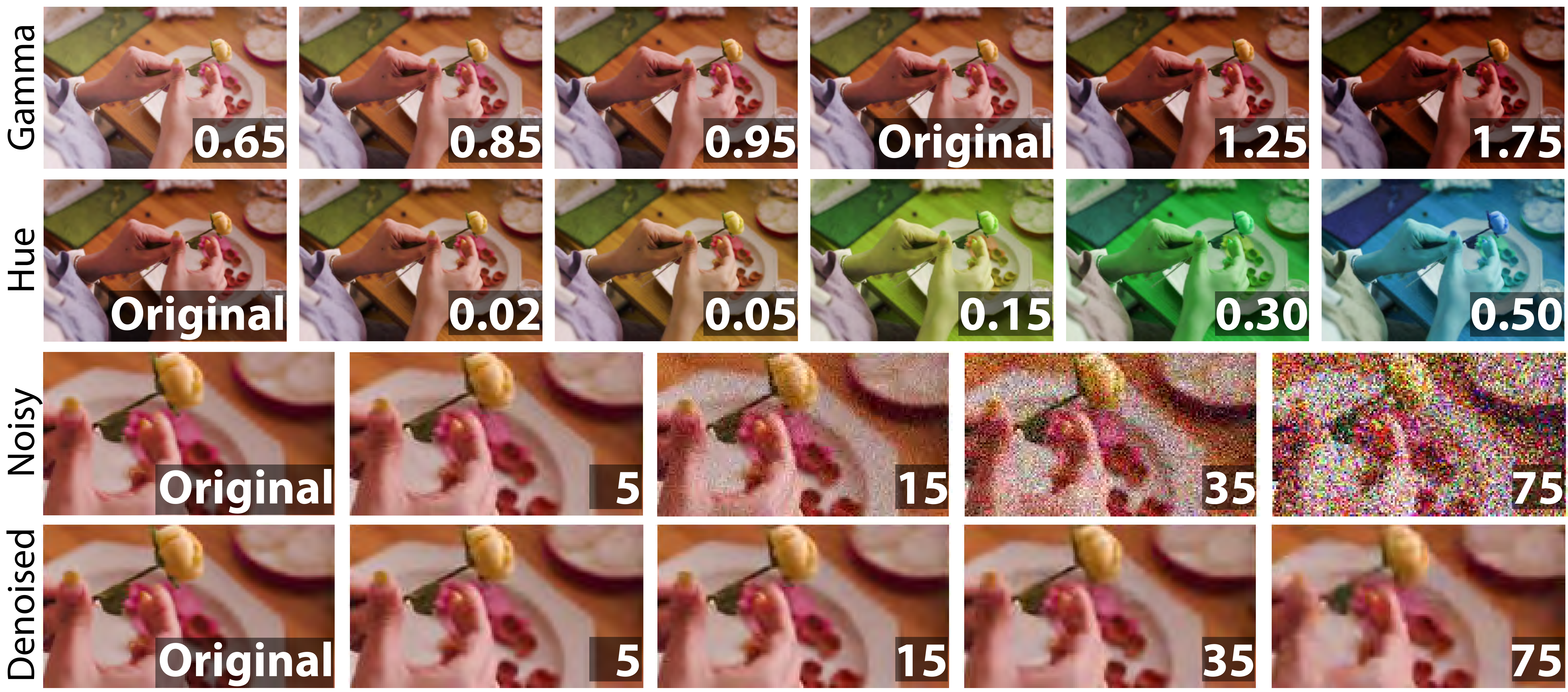}
  \vspace{-0.3in}
  \caption{\highlighttext{Example of auxiliary video inputs for the experiments in Sec.~\ref{sec:gammahue} (Gamma and Hue row) and Sec.~\ref{sec:noisy} (Noisy and Denoised row). The extent of perturbation is specified by the numbers in the lower right corner of images.}}
  \label{fig:ExpInp}
  \vspace{-0.17in}
\end{figure}

\highlighttext{Capturing high frame rate videos in low-light conditions leads to noisy auxiliary videos, which could potentially affect the performance of our system. To evaluate the effect of noise, we add zero-mean Gaussian noise with varying standard deviations to the auxiliary frames (see Fig.~\ref{fig:ExpInp}) and evaluate the quality of the estimated frames in Table~\ref{tab:AblationNoisy}. Our approach performs well at low levels of noise, but at high levels of noise, the performance deteriorates due to poor quality of optical flow estimation. We can mitigate this by denoising the auxiliary videos using an existing algorithm, such as VBM4D~\cite{vbm4d}, as shown in Table~\ref{tab:AblationNoisy}.}

\vspace{-0.07in}
\subsubsection{\highlighttext{Camera Desynchronization}}
\highlighttext{As previously discussed in Sec.~\ref{sec:RealCameras}, in addition to spatial misalignment, we also have temporal misalignment between the main and auxiliary videos. In this experiment, we deliberately desynchronize the auxiliary video from 0 to 3 frames and use main (1080p) and auxiliary (360p) videos with frame rates of 30 and 240 fps, respectively. As shown in Table~\ref{tab:AblationDesync}, our model's performance degrades gracefully with increase in desynchronization. Nevertheless, this is not a major challenge as the two cameras can be synchronized to trigger simultaneously on typical dual camera devices.}

\begin{table}[!t]
\renewcommand{\arraystretch}{1.3}
\caption{\highlighttext{Evaluating the effect of noise in auxiliary video on the result's quality. We show the results using both noisy and denoised  (using VBM4D~\cite{vbm4d}) auxiliary videos.}}
\vspace{-0.15in}
\centering
\begin{tabular}{l l c c c c c}
  \hline\hline
  Sigma & ( $\sigma$ ) & 0 & 5 & 15 & 35 & 75 \\ 
  \hline
   \multirow{2}{*}{Noisy} & LPIPS & \textbf{0.1005} & 0.1035 & 0.1076 & 0.1572 & 0.2747\\
   & SSIM & \textbf{0.906} & 0.904 & 0.898 & 0.821 & 0.689\\\cline{3-7}
   \multirow{2}{*}{Denoised} & LPIPS & \textbf{0.1005} & 0.1012 & 0.1036 & 0.1128 & 0.1380\\
   & SSIM & \textbf{0.906} & 0.905 & 0.903 & 0.896 & 0.869\\
  \hline\hline
\end{tabular}
\label{tab:AblationNoisy}
\vspace{-0.15in}
\end{table}

\begin{table}[!t]
\renewcommand{\arraystretch}{1.3}
\caption{\highlighttext{Evaluating the effect of temporal desynchronization between main and auxiliary videos}}
\vspace{-0.15in}
\centering
\begin{tabular}{l c c c c}
  \hline\hline
  \# frames & 0 & 1 & 2 & 3\\ 
  \hline
   LPIPS & \textbf{0.1005} & 0.1093 & 0.1213 & 0.1326\\
   SSIM & \textbf{0.906} & 0.888 & 0.875 & 0.862\\
  \hline\hline
\end{tabular}
\label{tab:AblationDesync}
\vspace{-0.17in}
\end{table}

\subsection{Limitations} 
\begin{figure}
  \includegraphics[width=\linewidth]{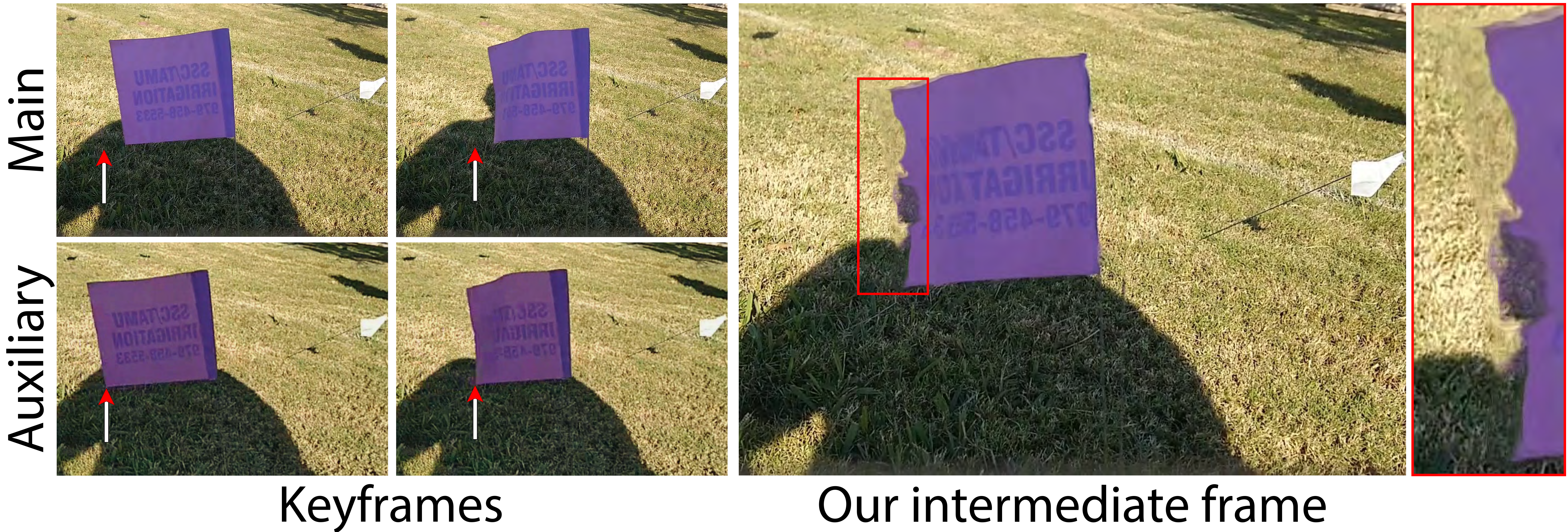}
  \vspace{-0.3in}
  \caption{\highlighttext{We show two consecutive main keyframes and their corresponding auxiliary keyframes. Here, since the flag is close to the camera rig, there is a large disparity between the main and auxiliary keyframes, as indicated by the arrows. Our flow estimation network cannot handle such a large misalignment and, thus, our system produces warping artifacts around the motion boundaries (see also supplementary video).}}
  \label{fig:LargeDisparity}
  \vspace{-0.1in}
\end{figure}

\begin{figure}
  \includegraphics[width=\linewidth]{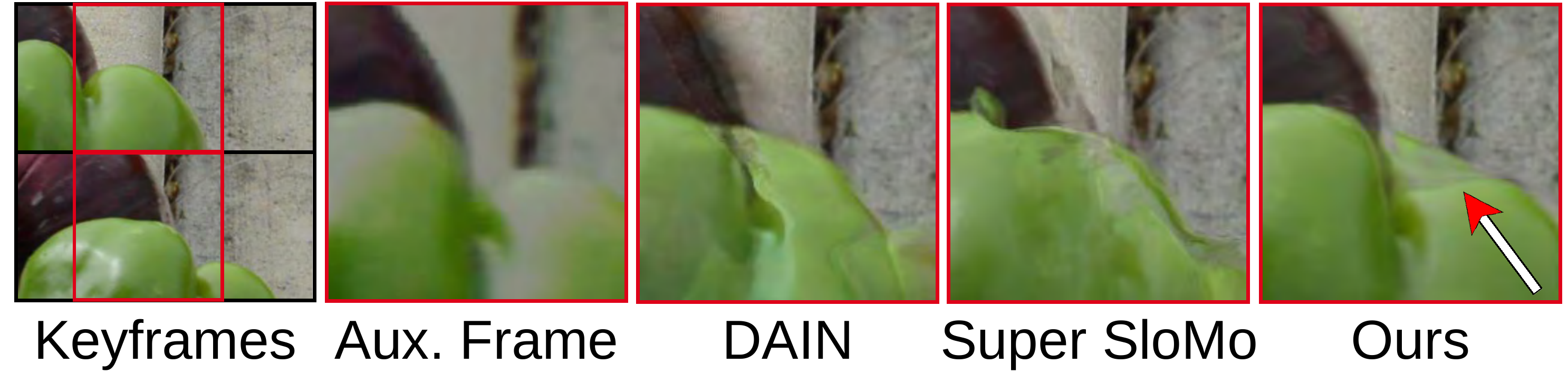}
  \vspace{-0.3in}
  \caption{Because of saturated regions in the auxiliary frame estimating a reliable flow is challenging. Although our result contains artifacts, it is considerably better than other methods.}
  \label{fig:Limitations}
  \vspace{-0.2in}
\end{figure}

Our approach has several limitations. First, our method is not able to utilize the information in the auxiliary frames, if they are captured at extremely low resolution (Table~\ref{tab:AblationPSNR}). Second, our system is unable to properly handle cases where the main and auxiliary videos have large baselines \highlighttext{or the object is very close to the rig causing significant misalignment (see Fig.~\ref{fig:LargeDisparity} and supplementary video)}. Third, in cases where the auxiliary video is saturated, it is difficult to estimate reliable flows. Therefore, our approach produces results with slight artifacts (Fig.~\ref{fig:Limitations}). However, our approach still generates better results than the existing methods. Finally, our approach is not able to produce results with higher frame rate than the auxiliary video. In the future, it would be interesting to extend our method to also interpolate the auxiliary frames.

%% file: Conclusions.tex
\section{Conclusion}
\label{sec:Conclusions}

We have presented a deep learning approach for video frame interpolation using a hybrid imaging system. In our method, we address the lack of temporal information in the input low frame rate video by coupling it with a high frame rate video with low spatial resolution. Our two-stage learning system properly combines the two videos by first aligning the high-resolution neighboring frames to the target frame and then combining the aligned images to reconstruct a high-quality frame. We generate the training data synthetically and perturb them to match the statistics of real data. We demonstrate that our approach outperforms prior work on both synthetic and real videos. We show the practicality of our approach on challenging scenes using real low-cost dual camera setups with small baseline.